\documentclass{article}
\usepackage{amssymb}
\usepackage{amsmath}
\usepackage{amsthm}
\usepackage[letterpaper]{geometry}
\usepackage[round]{natbib}
\usepackage[dvipsnames]{xcolor}
\usepackage[colorlinks,linktoc=true,
            citecolor= Cerulean!85!green!90!black,
            linkcolor=YellowOrange,
            urlcolor=RubineRed!85!black,
            anchorcolor=YellowOrange!85!black
            ]{hyperref}
            
\newif\ifdraft
\drafttrue

\usepackage{xparse}
\usepackage{xspace}
\usepackage[linesnumbered,ruled,vlined]{algorithm2e}
\usepackage{algorithm}
\usepackage{algpseudocode}
\SetKwInput{KwInput}{Input}                
\SetKwInput{KwOutput}{Output}              
\usepackage{fixltx2e}
\usepackage{pgfplots}
\usepackage{tikz}
\usepackage{float}
\usepackage{multirow}
\usepackage{multicol}
\usepackage{colortbl}
\usepackage{wrapfig}
\usepackage{array}
\usepackage{oplotsymbl}
\newcommand{\PreserveBackslash}[1]{\let\temp=\\#1\let\\=\temp}
\newcolumntype{C}[1]{>{\PreserveBackslash\centering}p{#1}}
\newcolumntype{R}[1]{>{\PreserveBackslash\raggedleft}p{#1}}
\newcolumntype{L}[1]{>{\PreserveBackslash\raggedright}p{#1}}
\usetikzlibrary{patterns}

\usepackage{graphicx}
\usepackage{mathtools}
\usepackage{float}
\usepackage{subcaption}
\usepackage{booktabs} 
\usepackage[shortlabels]{enumitem}

\usepackage{amssymb}
\usepackage{pifont}

\usepackage{bbold}

\usepackage{hyperref,amssymb,enumitem}

\setlist[itemize]{leftmargin=*}
\setlist[enumerate]{leftmargin=*}

\newcommand{\test}{\tilde{\bm{x}}}
\newcommand{\Test}{\tilde{\bm{X}}}
\newcommand{\layer}{\lambda}

\newcommand{\class}{y}

\newcommand{\bdist}{\bm{\mu}}
\newcommand{\vw}{\mathbf{w}}
\newcommand{\cmark}{\ding{51}}%
\newcommand{\xmark}{\ding{53}}%
\newcommand{\E}[1]{\operatorname{\mathbb{E}}[#1]}
\renewcommand{\P}[1]{\operatorname{\mathbb{P}}[#1]}
\newcommand*{\rej}{{\ooalign{\lower.3ex\hbox{$\sqcup$}\cr\raise.4ex\hbox{$\sqcap$}}}}

\usepackage{stackengine}
\renewcommand\underbar[1]{\stackunder[1pt]{$\operatorname{#1}$}{\rule{1.1ex}{.05ex}}}

\newcommand{\ind}[1]{\mathbb{1}_{\left[#1\right]}}

\renewcommand{\algorithmicrequire}{\textbf{Input: }}
\renewcommand{\algorithmicensure}{\textbf{Output: }}

\newcommand{\ie}{\textit{i.e.,}\@\xspace}
\newcommand{\eg}{\textit{e.g.,}\@\xspace}

\graphicspath{{images/}}

\ifdraft
\usepackage{todonotes}
\newcommand{\stephan}[1]{\todo[inline,color=green!40]{{\bf Stephan:~}#1}}
\newcommand{\adam}[1]{\todo[inline,color=blue!40]{{\bf Adam:~}#1}}
\newcommand{\mohammad}[1]{\todo[inline,color=orange!40]{{\bf Mohammad:~}#1}}
\newcommand{\nicolas}[1]{\todo[inline,color=red!40]{{\bf Nicolas:~}#1}}
\newcommand{\jonas}[1]{\todo[inline,color=teal!40]{{\bf Jonas:~}#1}}
\else 
\newcommand{\stephan}[1]{}
\newcommand{\adam}[1]{}
\newcommand{\mohammad}[1]{}
\newcommand{\nicolas}[1]{}
\newcommand{\jonas}[1]{}
\fi

\newcommand{\dknn}{DkNN\@\xspace}
\newcommand{\pdknn}{$p$-DkNN\@\xspace}
\newcommand{\pvalue}{$p$-value\@\xspace}
\newcommand{\pvalues}{$p$-values\@\xspace}

\newcommand{\mnist}{MNIST\@\xspace}
\newcommand{\fashion}{FashionMNIST\@\xspace}
\newcommand{\cifarX}{CIFAR10\@\xspace}

\newcommand{\rebel}{ReBeL\@\xspace}

\usepackage[utf8]{inputenc} 
\usepackage[T1]{fontenc}    
\usepackage{url}            
\usepackage{amsfonts}       
\usepackage{nicefrac}       
\usepackage{graphicx}
\usepackage{amsmath}
\usepackage{subcaption}
\usepackage{array}
\usepackage{multirow}
\usepackage{setspace}
\usepackage{booktabs}       
\usepackage{framed}
\usepackage{mathtools}
\usepackage{bm}
\usepackage{xspace}
\usepackage{todonotes}

\graphicspath{{./plots/}}

\DeclareMathOperator*{\argmax}{arg\,max}
\DeclareMathOperator*{\argmin}{arg\,min}

\title{
\pdknn: Out-of-Distribution Detection Through Statistical Testing of Deep Representations
}

\author{%
  Adam Dziedzic\thanks{Equal contribution.}\\
  University of Toronto \& Vector Institute\\
  \texttt{adam.dziedzic@utoronto.ca}\\
  \and
  Stephan Rabanser\footnotemark[1]\\
  University of Toronto \& Vector Institute \\
  \texttt{stephan@cs.toronto.edu}\\
  \and
  Mohammad Yaghini\footnotemark[1]\\
  University of Toronto \& Vector Institute\\
  \texttt{mohammad.yaghini@mail.utoronto.ca}\\
  \and
  Armin Ale\\
  University of Toronto \& Vector Institute\\
  \texttt{seyedarmin.alemohammad@mail.utoronto.ca}\\
  \and
  Murat A. Erdogdu\\
  University of Toronto \& Vector Institute\\
  \texttt{erdogdu@cs.toronto.edu}\\
  \and
  Nicolas Papernot\\
  University of Toronto \& Vector Institute\\
  \texttt{nicolas.papernot@utoronto.ca}\\
}

\date{} 

\begin{document}

\maketitle

\begin{abstract}
\noindent The lack of well-calibrated confidence estimates makes neural networks inadequate in safety-critical domains such as autonomous driving or healthcare. In these settings, having the ability to abstain from making a prediction on out-of-distribution (OOD) data can be as important as correctly classifying in-distribution data. We introduce \pdknn, a novel inference procedure that takes a trained deep neural network and analyzes the similarity structures of its intermediate hidden representations to compute \pvalues associated with the end-to-end model prediction. The intuition is that statistical tests performed on latent representations can serve not only as a classifier, but also offer a statistically well-founded estimation of uncertainty. \pdknn is scalable and leverages the composition of representations learned by hidden layers, which makes deep representation learning successful. Our theoretical analysis builds on Neyman-Pearson classification and connects it to recent advances in selective classification (\textit{reject option}). We demonstrate advantageous trade-offs between abstaining from predicting on OOD inputs and maintaining high accuracy on in-distribution inputs. We find that \pdknn forces adaptive attackers crafting adversarial examples, a form of worst-case OOD inputs, to introduce semantically meaningful changes to the inputs. 
\end{abstract}
\section{Introduction}

While deep representation learning can generalize well to in-distribution test data, the ability of neural networks to generalize---or even detect---out-of-distribution data remains weak~\citep{nagarajan2021understandingOOD}. This is an obstacle to their deployment in tasks which raise concerns of safety (e.g., autonomous driving~\citep{jain2019analyzing, ghodsi2021generating, wang2021advsim}), security (e.g., malware detection~\citep{anderson2017evading}), or ethics (e.g., fairness of lending decisions~\citep{liu2018delayed, chen2018fair}). Certain approaches to machine learning promise to address this limitation by inherently modeling uncertainty. This is best exemplified by Bayesian neural networks, which model the distribution of each weight in the neural network to naturally obtain a distribution of outputs from which uncertainty can be estimated. Unfortunately, this class of approaches often lacks scalability~\citep{gal2014distributed, gustafsson2020evaluating}.

In this paper, we instead reason about uncertainty in the paradigm of frequentist statistics. The crux of our approach relies on the observation that a statistical test can be a natural classifier. Furthermore, this classifier comes with the added benefit of modeling type I error, i.e., the ability to abstain from making a prediction when significance is too low. It is thus a compelling option to both model uncertainty and detection of out-of-distribution samples fed into the classifier.

To combine the advantages of this frequentist approach with those of deep representation learning, we propose to leave the training algorithm for deep neural networks (DNNs) intact but modify how one uses a trained model for inference. Our prediction procedure exploits the modularity of DNNs: each hidden layer outputs a different learned representation whose similarity structure helps us characterize the data distribution of interest. We analyze these similarity structures to compute a \pvalue associated with accepting each possible label. Our analysis shares some intuition with the Deep k-Nearest Neighbor \citep{dknn}, hereafter abbreviated as \dknn, where a nearest neighbor search in representation spaces learned by each of the hidden layers serves as the basis for predicting on a test input---in lieu of directly outputting the result of a forward pass through the neural network. While \citet{dknn} reported promising initial results for out-of-distribution detection, including adversarial examples~\citep{biggio2013adv,szegedy2014adv}, \dknn's uncertainty estimation is not statistically well-founded. 

\begin{wrapfigure}{r}{0.5\linewidth}
	\vspace{-6pt}
	\scalebox{0.5}{
		\pgfmathdeclarefunction{gauss}{2}{%
	\pgfmathparse{1/(#2*sqrt(2*pi))*exp(-((x-#1)^2)/(2*#2^2))}%
}
\begin{tabular}{c}
\begin{tikzpicture}
	    \tikzstyle{every node}=[font=\LARGE]

	\begin{axis}[
		domain=-5:5, samples=100,
		axis y line=none,
		axis x line=bottom,
		every axis y label/.style={at=(current axis.above origin),anchor=south},
		every axis x label/.style={at=(current axis.right of origin),anchor=west},
		height=5cm, width=15cm,
		xtick={-2, 0, 2}, 
		ytick=\empty,
		enlargelimits=false, 
		clip=false, 
		axis on top,
		title={(1)}
		]
		\addplot [densely dashdotted] {gauss(-2,1)} 
		[yshift=8pt] node [pos=0.3] {OOD}
		[yshift=30pt] node [pos=0.1, text width=3.5cm, align=left] {Classical Classification};
		\addplot [very thick] {gauss(2,1)} 
		[yshift=8pt] node [pos=0.70] {ID};
		\addplot [fill=red!20, draw=none, domain=-4:0] {gauss(2,1)}
		node [pos=0.96, yshift=-5pt] {II}
		\closedcycle
		;
		\addplot [fill=cyan!20, draw=none, domain=0:5] {gauss(-2,1)}
		node [pos=0.04, yshift=-5pt] {I}
	 \closedcycle;
 		\addplot[dashed] coordinates {(0,0) (0,{gauss(-2,1)})};

	\end{axis}		
\end{tikzpicture}
\\
\begin{tikzpicture}
	\tikzstyle{every node}=[font=\LARGE]
\begin{axis}[
	domain=-5:5, samples=100,
	axis y line=none,
	axis x line=bottom,
	every axis y label/.style={at=(current axis.above origin),anchor=south},
	every axis x label/.style={at=(current axis.right of origin),anchor=west},
	height=5cm, width=15cm,
	xtick={-2, 0.35, 2}, 
	ytick=\empty,
	enlargelimits=false, 
	clip=false, 
	axis on top,
	title={(2)}
	]
	
	\addplot [densely dashdotted] {gauss(-2,1)}
	[yshift=30pt] node [pos=0.1, text width=3cm, align=left] {NP-Classification};
	\addplot [very thick] {gauss(2,1)};
	\addplot [fill=red!20, draw=none, domain=-5:0.35] {gauss(2,1)} 
	\closedcycle;
	\addplot [fill=cyan!20, draw=none, domain=0.35:5] {gauss(-2,1)}
 \closedcycle;

	\addplot[dashed] coordinates {(0.35,0) (0.35,{gauss(2,1)})};
\end{axis}	
\end{tikzpicture}
\\
\begin{tikzpicture}
	\tikzstyle{every node}=[font=\LARGE]
	\begin{axis}[
		domain=-5:5, samples=100,
		axis y line=none,
		axis x line=bottom,
		every axis y label/.style={at=(current axis.above origin),anchor=south},
		every axis x label/.style={at=(current axis.right of origin),anchor=west},
		height=5cm, width=15cm,
		xtick={-2, -0.5, 0.35, 2}, 
		xticklabel style={rotate=45},
		ytick=\empty,
		enlargelimits=false, 
		clip=false, 
		axis on top,
		title={(3)}
		]
		
		\addplot [very thick] {gauss(2,1)};
		\addplot [densely dashdotted] {gauss(-2,1)}
		[yshift=30pt] node [pos=0.1, text width=3.5cm, align=left] {Classical Classification};
		\addplot [dotted] {gauss(1.6, 1)} [yshift=20pt] node [pos=0.63] {Shifted};
		\addplot [dotted] {gauss(1.3, 1)};
		\addplot [dotted] {gauss(1, 1)};
		\addplot [fill=red!20, opacity=0.2, draw=none, domain=-4:0] {gauss(2,1)} 
		\closedcycle;
		\addplot [fill=cyan!20, opacity=0.2, draw=none, domain=0:4] {gauss(-2,1)}
		\closedcycle;
		\addplot [fill=red!20, opacity=0.2,draw=none, domain=-4:-0.2] {gauss(1.6,1)} 
		\closedcycle;
		\addplot [fill=cyan!20, opacity=0.2, draw=none, domain=-0.2:4] {gauss(-2,1)}
		\closedcycle;
		\addplot [fill=red!20, opacity=0.2,draw=none, domain=-4:-0.35] {gauss(1.3,1)} 
		\closedcycle;
		\addplot [fill=cyan!20, opacity=0.2, draw=none, domain=-0.35:4] {gauss(-2,1)}
		\closedcycle;
		\addplot [fill=red!20, opacity=0.2,draw=none, domain=-4:-0.5] {gauss(1,1)} 
		\closedcycle;
		\addplot [fill=cyan!20, opacity=0.2, draw=none, domain=-0.5:4] {gauss(-2,1)}
		\closedcycle;

		\addplot[dashed] coordinates {(0,0) (0,{gauss(2,1)})};
		\addplot[dashed] coordinates {(-0.2,0) (-0.2,{gauss(1.6,1)})};
		\addplot[dashed] coordinates {(-0.35,0) (-0.35,{gauss(1.3,1)})};
		\addplot[dashed] coordinates {(-0.5,0) (-0.5,{gauss(1,1)})};
				
	\end{axis}	

\end{tikzpicture}
\\
\begin{tikzpicture}
	\tikzstyle{every node}=[font=\LARGE]
	\begin{axis}[
		domain=-5:5, samples=100,
		axis y line=none,
		axis x line=bottom,
		every axis y label/.style={at=(current axis.above origin),anchor=south},
		every axis x label/.style={at=(current axis.right of origin),anchor=west},
		height=5cm, width=15cm,
		xtick={-2, 0.35, 2}, 
		ytick=\empty,
		enlargelimits=false, 
		clip=false, 
		axis on top,
		title={(4)}
		]
		
		\addplot [very thick] {gauss(2,1)};
		\addplot [densely dashdotted] {gauss(-2,1)}
			[yshift=30pt] node [pos=0.1, text width=3cm, align=left] {NP-Classification};
		\addplot [dotted] {gauss(1.6, 1)};
		\addplot [dotted] {gauss(1.3, 1)};
		\addplot [dotted] {gauss(1, 1)};
		\addplot [fill=red!20, opacity=0.2, draw=none, domain=-4:0.35] {gauss(2,1)} 
		\closedcycle;
		\addplot [fill=red!20, opacity=0.2,draw=none, domain=-4:0.35] {gauss(1.6,1)} 
		\closedcycle;
		\addplot [fill=red!20, opacity=0.2,draw=none, domain=-4:0.35] {gauss(1.3,1)} 
		\closedcycle;
		\addplot [fill=red!20, opacity=0.2,draw=none, domain=-4:0.35] {gauss(1,1)} 
		\closedcycle;
		\addplot [fill=cyan!20, draw=none, domain=0.35:4] {gauss(-2,1)}
		\closedcycle;
		
		\addplot[densely dashdotted] coordinates {(0.35,0) (0.35,{gauss(1,1)})};
	\end{axis}	
	
\end{tikzpicture}	
\end{tabular}
	}
\caption{Selection of in-distribution samples using classical  (1) vs. NP (2) classification  where the Type I error $< 0.05$. (3) and (4) repeat this in the presence of distribution shifts. An NP-classifier gives a Type I error guarantee at the expense of higher Type II error.}
 \vspace{-0.2in}
\label{fig:np-vs-normal}
\end{wrapfigure}
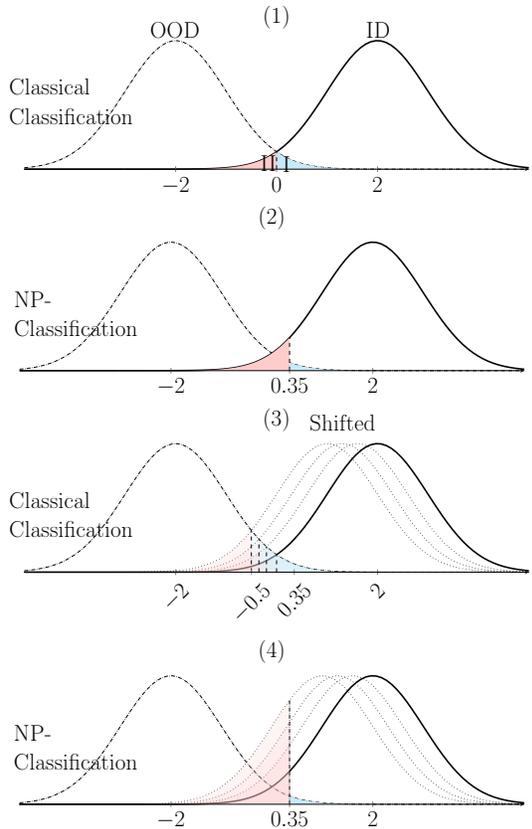

To remediate these pitfalls, we propose a novel inference procedure called \pdknn, which computes proper \pvalues. At inference time, \pvalues are calculated through pairwise testing of mean distances between the test input and k-closest neighbors in each class and layer. Similar approaches also based on statistical hypothesis testing of nearest neighbor latent representations have recently been proposed by~\citet{raghuram2020detecting}. While strongly related, their work introduces new test statistic based on log-likelihood-ratios between nearest-neighbor class counts. Furthermore, they make the assumption that  tests across layers are independent to aggregate them. Our approach on the other hand uses proven off-the-shelf testing statistics and employs aggregation methods that do not impose any independence assumptions.
We combine \pvalues using known statistical techniques to obtain a meaningful single metric of uncertainty for the entire model. 

Our statistical testing evaluates the probability of each class regardless of whether this is the correct class or not, whereas prior approaches evaluated the probability of a given class assuming that it was the correct class. 

\pdknn is a skeptical/cautious method by design. Abstaining is the default action since that is our null hypothesis throughout the tests.

This has advantages which are not without raising new difficulties when it comes with defining and evaluating what constitute good performance from the classifier. 
 In the majority of the ML literature the focus is on minimizing the risk, which is a weighted sum of false positive and false negative rates. 
Instead, robust classification is a multi-objective optimization problem. Robust classifiers require us to step away from the assumption that sample distributions are the same (or even close) at train and inference time. As a result, a robust classifier has two (possibly competing) objectives: \textbf{(i)} selecting the correct samples to classify (the selection step, or \textit{S-step}),  and \textbf{(ii)} classifying the selected samples accurately (classification step, or \textit{C-step}). Consequently, the performance metrics are also different for each goal: for the latter, we want to have the best accuracy, whereas for the former, it is more important for a robust classifier to abstain from selecting incorrect samples (small Type I error / low false positive rate) than to erroneously discard some correct ones: e.g., classifying a malicious tumor as benign is a deadly mistake compared to misclassifying a benign tumor as malicious.

The Neyman-Pearson (NP-) classification is a more appropriate framework than empirical risk minimization since it bounds the most dangerous error (typically called Type I error) at the expense of the less significant error (Type II). We posit that for safety or security critical classifiers that may be provided arbitrary samples at inference time, a selection step should precede the actual classification; and that NP-classification is an appropriate framework for that.

Our empirical evaluation of \pdknn in Section~\ref{sec:experiments} commences with the calibration of our predictor. This includes tuning the number of nearest neighbors and search for other hyperparameters to optimally compute and aggregate \pvalues, for example, we adjust the significance level. Then, we evaluate the \pdknn on standard vision datasets (\mnist, \fashion, SVHN, GTSRB, and \cifarX) to show that (1) it is able to reject out-of-distribution inputs while continuing to correctly classify in-distribution inputs; (2) it offers better trade-offs between accuracy and false positives on legitimate data than prior approaches including \dknn or Trust~\citep{jiang2018trust}; and (3) it also offers some robustness to attackers crafting worst-case out-of-distribution inputs such as (adaptive) adversarial examples. We note that for (1) and (2) we focus on a safety scenario where out-of-distribution data is \textit{natural} (e.g., from a different dataset of the same modality) whereas in (3) we switch to a security-motivated setting where out-of-distribution examples are \textit{crafted} by an adversary. Thus, security implies safety, but the converse is not true.

To summarize, our contributions are as follows: 
\begin{itemize}
    \item We introduce the \pdknn inference procedure which takes in a trained DNN and incorporates validation data at test time to output \pvalues and effect sizes capturing prediction uncertainty.
    \item We theoretically analyze properties of \pdknn to
    establish a connection to recent advances in selective classification. In particular, we show that \pdknn is a PQ-Learner~\citep{goldwasser2020beyond}. 
    \item We empirically demonstrate that \pdknn abstains from predicting on out-of-distribution inputs while maintaining high accuracy on in-distribution inputs. 
    In the presence of an adversary aware of \pdknn, we show that adversarial examples able to evade \pdknn are more likely to introduce semantically meaningful changes to the input---thus defeating the purpose of the attack.
\end{itemize}
\section{Related Work}
\label{sec:related-work}
\paragraph{Uncertainty Quantification in Deep Learning.}
High-stakes decision-making domains call for proper uncertainty estimates along with  point prediction such that risk can be effectively minimized \citep{gal2016uncertainty}. 
From a Bayesian viewpoint, Gaussian processes~\citep{rasmussen2003gaussian} offer a principled way of quantifying uncertainty by modeling the probability distribution over all possible functions that fit the data. Although mathematically elegant, Gaussian processes usually fail at scaling to  complex data modeling tasks. As an alternative to Gaussian processes, Bayesian neural networks offer a natural way of accounting for full predictive uncertainty by estimating a probability distribution over the model's parameters. In order to be practically feasible, techniques from variational inference~\citep{blundell2015weight} or Markov chain Monte Carlo (MCMC)~\citep{mcmc4ml} are used to compute an approximate posterior over the model's parameters. To further reduce the computational cost, Monte Carlo dropout~\citep{gal2016dropout} can be applied to yield predictive uncertainty measures by performing dropout at test time. When taking a frequentist viewpoint, ensembling techniques~\citep{lakshminarayanan2016simple} combine multiple models that are either trained on different subsets of the training data or trained using different weight initializations. 
Finally, our work is perhaps closest to the Deep k-Nearest Neighbors (DkNN)~\citep{dknn} which we covered in the introduction. 

\paragraph{OOD, Safety, and Security.} In our work, we consider OOD detection in the context of both safety and security. By safety, we mean settings where the model may be presented with naturally occurring OOD inputs~\citep{amodei2016concrete}. This is the case for naturally occuring transformations of the input, which can prove to be challenging even for state-of-the-art vision models~\citep{ford2019adversarial,hendrycks2018benchmarking}. Instead, when we consider an adversary attacking our model, its security will stem from the ability to detect malicious OOD inputs such as adversarial examples~\citep{biggio2013adv,szegedy2014adv} or recover the correct labels for them.
Many recent lines of works studied different causes of adversarial examples, postulating linearity and over-parametrization as possible culprits~\citep{biggio2013adv,goodfellow2014explaining}. An intuitive explanation is that adversarial examples are caused by small measure regions of adversarial class \textit{jutting} into a correct decision region~\citep{roth19a}. This model is commonly accepted since the clean in-distribution samples remain correctly classified when perturbed with small random noise (e.g., Uniform or Gaussian).
Adversarial examples directly exploit the poor generalization of neural networks and this makes them difficult to detect or correct their predictions.  
We experiment with standard attacks, such as FGSM and PGD, as well as adaptive white-box attacks, for example, the feature adversaries introduced by~\citet{feature2016attack}. 

\section{Methods}
\label{sec:method}

\begin{figure*}
    \centering
    \includegraphics[width=\linewidth]{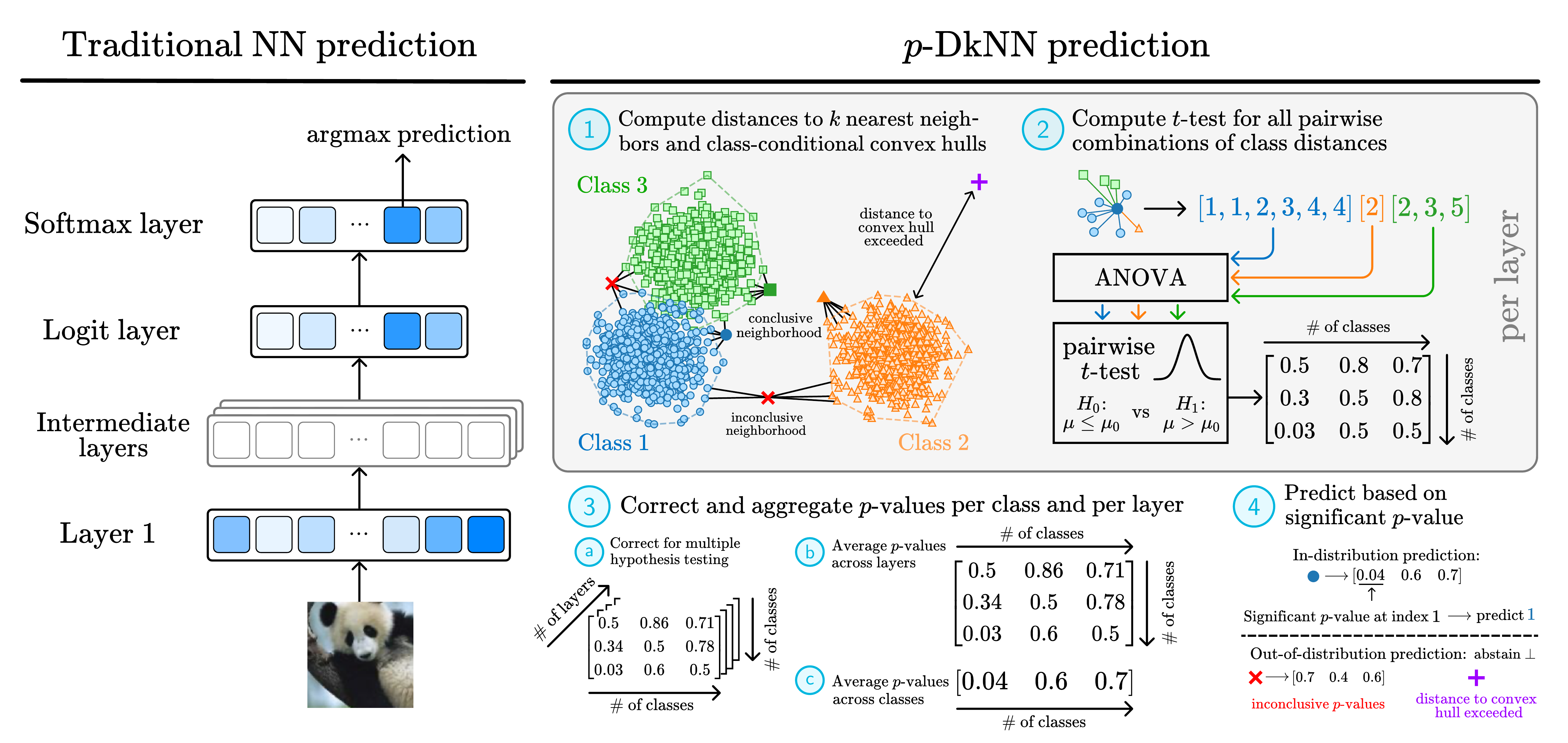}
    \caption{\textbf{Overview of \pdknn vs traditional DNN classification}. For a given test input, we compute the representations across all latent layers and estimate the nearest neighbor distances with respect to activations of a validation set. After passing an ANOVA test, these distances are fed into a sequence of pairwise $t$-tests. The resulting \pvalue tensor is then corrected for multiple hypothesis testing and aggregated to a single vector of class-\pvalues. Significant \pvalues yield a class prediction, otherwise we abstain. We also abstain 
    if a data point is too far from a predicted class' convex hull. Effect sizes are omitted for clarity.}
    \label{fig:overview}
\end{figure*}

We propose \pdknn which takes a DNN trained using an off-the-shelf optimizer and adapts its inference procedure to harness statistical hypothesis testing of intermediate layer representations. 
Rather than 
taking the argmax of the softmax output layer, \pdknn performs prediction by computing distances between internal representations of a given test point and the representations of all points from a validation set $\bm{X} \in \mathbb{R}^{V \times D}$, for which label information $\bm{y} \in [C]^V$ is available, across all layers and feeding these distances into a sequence of statistical hypothesis tests and subsequent aggregation methods\footnote{To aid generalization, we fit the underlying neural network classifier using \emph{training} data while \pdknn's inference procedure relies on computing distances to a \emph{validation} set.}. We depict \pdknn's prediction stage in Figure \ref{fig:overview} and provide an  overview in Algorithm \ref{alg:pdknn}. 

\subsection{Distance computation}

For a given test input $\test \in \mathbb{R}^D$, \pdknn starts by computing the activations $\{\bm{\tilde{x}}_{\lambda}\}_{\lambda=1}^{L}$ of $\test$ for all layers $\layer$ by performing a forward pass through the network. Then, for each layer $\layer$, we compute the $\ell_2$ distances between the test point's representation at layer $\layer$, i.e. $\bm{\tilde{x}}_{\lambda}$, and the representations of all data points from a validation set $\{\bm{X}_{\lambda}\}_{\lambda=1}^{L}$. We collect these distances in $\bdist \in \mathbb{R}^V$ with entries $\bdist_j = ||\bm{X}_{\lambda,j} - \bm{\tilde{x}}_\layer||_2$.  
To estimate the distances and labels of the validation points in the neighborhood of $\bm{\tilde{x}}$ only, 
we select the first $k$ elements of $\bdist$ in ascending order. We store these $k$-nearest-neighbor distances and their corresponding ground truth labels in $\hat{\bdist} \in \mathbb{R}^k$ and $\hat{\bm{y}} \in [C]^k$, respectively.

\subsection{Classification by means of hypothesis testing}
Having obtained $\hat{\bdist}$ and $\hat{\bm{y}}$ at a particular layer $\layer$, we want to determine whether there is sufficient evidence encoded in the distances such that the test point $\bm{\tilde{x}}$ can be confidently assigned to a particular class. Since we approach this problem through the lens of statistical testing, we first need to (i) determine whether there is any difference in the distances from the classes present in the local neighborhood; and if such a difference exists (ii) investigate the distances pairwise between classes to determine the dominating class(es). 

For (i), we perform a one-way Welch ANOVA study with null hypothesis $H_0: \hat{\bdist}_{\hat{y}=1} = \hat{\bdist}_{\hat{y}=2} = \dots =  \hat{\bdist}_{\hat{y}=C}$ at significance level $\alpha$, which if rejected, proves that there is at least one class that is on average closer (or farther) to $\test$ at layer $\layer$. Here $\hat{\bdist}_{\hat{y}=i}$ are the distances to the closest validation points from class $i$. 
For (ii), i.e. to find which class is the closest and by how much, we must perform additional post-hoc tests. We propose to employ a pair-wise $t$-testing setup between all the distances from the classes present in $\hat{\bm{y}}$. If $C'$-many distinct classes are present in $\hat{\bm{y}}$, then we carry out $C' \choose 2$ pairwise comparisons between class distances. Given a distance vector from class $c_1$, $\hat{\bdist}_{\hat{y}=c_1}$, and a distance vector from class $c_2$, $\hat{\bdist}_{\hat{y}=c_2}$, we set up a null hypothesis $H_0: \hat{\bdist}_{\hat{y}=c_1} \leq \hat{\bdist}_{\hat{y}=c_2}$ stating that the mean class distance of class $c_2$ is greater or equal than the mean class distance of class $c_1$. We also establish the corresponding alternative $H_1: \hat{\bdist}_{\hat{y}=c_1} > \hat{\bdist}_{\hat{y}=c_2}$ under which the mean class distance of class $c_2$ is strictly smaller than the mean class distance of class $c_1$. A significant result therefore implies that the test data point $\tilde{\bm{x}}$ is closer to $c_2$ than to $c_1$. Carrying out this testing procedure for all pair-wise combination of classes yields a matrix $\bm{P} \in [0,1]^{C \times C}$ containing all \pvalues  for the respective pair-wise tests.

In addition to \pvalues, the performed $t$-tests also returns an effect size for all combinations of classes tested. For a class pair $c_1$, $c_2$ the effect size quantifies the magnitude of the difference between two groups and is defined as $e = \frac{\hat{\bdist}_{\hat{y}=c_1} - \hat{\bdist}_{\hat{y}=c_2}}{\sigma}$ where $\sigma$ is estimated from $\hat{\bdist}_{\hat{y}=c_2}$. Similar as with \pvalues, we collect the effect sizes in a matrix $\bm{E} \in \mathbb{R}^{C \times C}$.

 \begin{wrapfigure}{R}{0.5\textwidth}
 \hspace{15pt}
 \begin{minipage}{0.45\textwidth}
     \vspace{-10pt}
     \input{contents/pdknn_alg}
     \vspace{-0.3in}
 \end{minipage}
 \end{wrapfigure}
 
\subsection{Aggregation of p-values}

Performing the pair-wise statistical hypothesis testing procedure for each layer yields a collection of matrices $\{\bm{P}_\lambda\}_{\lambda = 1}^L$, $\{\bm{E}_\lambda\}_{\lambda = 1}^L$ that can be combined into tensors $\mathcal{P}$, $\mathcal{E}$ for the \pvalue and effect sizes, respectively. Our final goal being to obtain a vector of $p$-values
signaling the support strength for 
each class, we devise an aggregation scheme through multiple hypothesis testing correction and averaging.

First, we need to ensure that our false positive rate is not inflated due to carrying out multiple tests on the same data. Thus, we apply a two stage false discovery rate correction on top of the \pvalue tensor $\mathcal{P}$ which enables us to bound incorrect rejections of the null.

Next, we start aggregating \pvalues across layers, i.e. we compute a single \pvalue for a fixed combination of classes $c_1$ and $c_2$ based on the full \pvalue tensor $\mathcal{P}$. We do that by calculating the arithmetic mean over the collection of \pvalues between class $c_1$ and $c_2$ and multiplying the resulting mean by 2, or for a weighted average with weights $\bm{w}$, the multiplicative factor is $\min(2, \frac{1}{\max(\bm{w})})$. If an averaged \pvalue is larger than $1$ after aggregation, we clip it to $1$. This particular choice of aggregation is supported by~\citet{vovk2019combining}, which states that \pvalues can be averaged independent of their dependence structure.
While~\citet{vovk2019combining} presents a general framework supporting multiple different mean computations, 
the (weighted) arithmetic averaging method is preferred in cases of strong \pvalue dependence (which we typically expect between latent neural network representations).

Having now obtained a matrix of \pvalues, we still have to aggregate across one of the classes dimensions. Since our hypothesis testing setup yields significant results if the column class has smaller mean distance, we aggregate across columns instead of rows. The aggregation is performed identical to the previous stage: we compute the arithmetic mean across columns and multiply the result by 2. 
In addition to $\mathcal{P}$, we also aggregate the effect sizes $\mathcal{E}$ into a single vector. Since effect sizes do not need to abide by strict aggregation rules dictating the consistency of the final aggregation (as was the case with \pvalues), we simply compute the arithmetic mean across layers and class columns without needing to apply any correction factor.

\subsection{Prediction based on significant \pvalue}

The final aggregation result is a vector of \pvalues $\bm{p} \in [0,1]^{C}$ and a vector of effect sizes $\bm{e} \in \mathbb{R}^{C}$. To make a classification decision for $\test$, we query $\bm{p}$ for the minimum \pvalue $\tilde{p} = \min \bm{p}$ and the index of the maximum effect size $\tilde{y} = \argmax \bm{e}$. If the minimum \pvalue $\tilde{p}$ happens to be significant (i.e below a chosen significance level $\alpha$), then we return the index $\tilde{y}$ as the classification result. However, if no single \pvalue in $\bm{c}$ happens to be significant, then we abstain  (i.e. return  $\bot$). Note that \pvalues are being used to determine whether we should reject or accept $\test$, while the effect sizes determine the classification result for an accepted test sample. This separation of concerns allows us to also make a confident classification decision in case multiple \pvalues in $\bm{p}$ fall below our significance level $\alpha$.

\subsection{Distance computation to convex hull}

The proposed method allows us to correctly classify inputs that are supported by sufficient evidence from one specific class $\tilde{y}$ or to abstain from prediction in case there is a large amount of disagreement in the local neighborhood around the test input $\test$. However, some OOD data points might lie far from the density defined by the validation points $\{\bm{X}_{\lambda}\}_{\lambda=1}^{L}$ but at the same time be closest to one particular class. To detect this type of OOD data as well, we propose to additionally validate the distance of a test input to the predicted class $\tilde{y}$. Concretely, we compute the convex hull of the predicted class and reject a data point if its distance from the convex hull exceeds a pre-calibrated maximum threshold $\gamma$.

\section{Theoretical Connections \& Analysis}
\label{sec:theory}

\begin{table}
    \caption{\textbf{Confusion Matrix for the Extended Binary Classification Scenario.} \textbf{S}election, \textbf{C}lassification, and Type \textbf{I} and \textbf{II} errors are marked.}
    \label{tab:extended_confusion}
    \centering
    \vspace{10pt}
    \begin{tabular}{L{2mm} L{2mm} C{15mm} C{15mm} C{5mm} C{5mm}}
        \toprule
         \multicolumn{2}{c}{Dec \textbackslash True}  & $\top, 0$ & $\top, 1$ & $\bot, 0$ & $\bot, 1$\\
        \midrule
         $\top$ & 0 & \cmark & \xmark$\;$(\textbf{C-II}) & \multicolumn{2}{c}{\multirow{2}{*}{\textbf{S-I}}}\\
         $\top$ & 1 & \xmark$\;$(\textbf{C-I}) & \cmark& &\\
         $\bot$ & 0 & \multicolumn{2}{c}{\multirow{2}{*}{\textbf{S-II}}}  & \multicolumn{2}{c}{\multirow{2}{*}{\LARGE\cmark}} \\
         $\bot$ & 1 & & & &\\ 
     \bottomrule
    \end{tabular}
\end{table}
We pursue two goals in our extended classification scenario. First, we want to make sure that we do not produce overconfident classification decisions. We call this step the \textbf{S}election-step, or S-step for short. The S-step helps us avoid OOD and adversarial sample classification. Second, we want to make sure we produce accurate classification decisions for samples that we accept in the S-step. We denote this as the \textbf{C}lassification step, or C-step. The C-step is akin to empirical loss minimization, as is common in statistical learning. The S-step is always a binary decision, regardless of the number of classes in the target problem (C-step). To simplify our theoretical formulation and analysis, we consider a binary classification scenario for the C-step, too. Table \ref{tab:extended_confusion} shows the confusion table for our extended setup.
Due to the nature of our problem setup, the S-step conceptually precedes the C-step, \ie if a sample is OOD without a doubt (\eg from the preceding $S$ distribution) then classifying it would produce a vacuous decision regardless of the fact that the calculated label matches a ground-truth label, or not.

\paragraph{NP-Classification and NP-ERM Formulation} 
In the classification step, we care most about the classification accuracy; or minimizing the risk $R(h) = \E{\ind{h(x) \neq y}} = \P{h(x) \neq y}$ which is the weighted sum of type I and type II errors: $R(h) = \P{Y=0}R_0(h) + \P{Y=1}R_1(h)$. As we previously discussed, for a safety-critical classifier, in the selection step, it is essential that the type I error (that is, probability of classifying OOD samples as ID) be bounded and thus a NP-Classification scenario is more appropriate. Therefore, we formulate our setup in an NP-ERM setting~\citep{scott_neyman-pearson_2005}:

Let $h \in \mathcal{H}$ be a trained model hypothesis, $x \in \mathcal{X} \subseteq \mathbb{R}^d$ our test input, $\mathcal{Y} = \{0,1\}$ the space of C-step ground truth labels, and $\mathcal{Z} = \{\bot,\top\}$ the space of S-step ground truth labels. We define $f(x) = {(m(x), h(x))} \rightarrow  \{\bot, \top\} \times \{0,1\}$ as the extended classifier's decision function which uses the aggregated \pvalues for class $c \in \mathcal{Y}$: $\underbar p_c(x) = (p_{c, 1}, \dots, p_{c,L}) ^ T$ to determine the the selected samples. The corresponding effect sizes $\underbar{es}_c(x) = (es_{c, 1}, \dots, es_{c,L}) ^ T$ are used in the classification step. We formulate our approach as the following optimization problem:
\begin{align}
    \begin{split}
    \underset{\vw \in \Delta(L), \; \omega \in [0, 1]}{\argmin}\quad & \frac{\omega}{n}
    \sum_{x \in \mathcal{X}} \ind{m(x) = \bot, z=\top}\ + \\
    & \frac{1 - \omega}{|\{m(x) = \top\}|}
    \sum_{m(x) = \top} \ind{h(x) \neq y } \label{eq:loss}
    \end{split}
\end{align}
such that
\begin{align}
  & \frac{1}{n}\sum_{x \in \mathcal{X}}\ind{m(x) = \top, z=\bot} < \alpha + \epsilon \label{eq:s-type-1}\\
& \frac{1}{|\{m(x) = \top\}|}\sum_{m(x) = \top} \ind{h(x) = 1, y = 0} < \alpha + \epsilon \label{eq:c-type-1}\\
& m(x) = \begin{cases}
\top &\langle \mathbf{w}, \underbar{p}_0(x) \rangle < \alpha + \epsilon \;\vee\; \\
& \langle \mathbf{w}, \underbar{p}_1(x) \rangle < \alpha + \epsilon\\
\bot &\text{otherwise}
\end{cases} \label{eq:s-step}\\
& h(x) = \argmax_{c \in \mathcal{Y}=\{0, 1\}} \quad {\langle \mathbf{w}, \underbar{es}_c(x) \rangle} \label{eq:c-step}
\end{align}

In \eqref{eq:loss}, we are minimizing the weighted loss of type II error in the S-step (\ie rejecting an ID sample) and the misclassification of the C-step. Weights $\vw$ are on the $L$-dimensional probability simplex $\Delta(L)$. $\omega \in [0, 1]$ is a hyper-parameter that helps us tune this trade-off.  Since we are operating in an NP hypothesis testing framework, both the S-step and the C-step type I errors are bounded per \eqref{eq:s-type-1} and \eqref{eq:c-type-1} with slack parameter $\epsilon$. Equation~\eqref{eq:s-step} represents the selection step via the meta-hypothesis test $\langle \mathbf{w}, \underbar{p} \rangle < \alpha$, where $\mathbf{w}$ are the weights assigned to each layer. Finally, \eqref{eq:c-step} chooses the class with largest effect size (C-step). It is possible that after the S-step, the accepted samples show exclusive evidence for only one class; in that case, the C-step and \eqref{eq:c-step} are trivial.

\paragraph{What constitutes OOD \& Connection to PQ-Learning~\citep{goldwasser2020beyond}}

\begin{wrapfigure}{r}{0.5\linewidth} 
	\hspace{5pt}
%
	\begin{tikzpicture}
		\draw[pattern=north east lines] (0,0) rectangle (7.5,5);  
		\node [fill=white, at={(.6,.3)}, rounded corners] {$OOD$};
		
		\draw [fill, color=white, draw=black](3.2,2.5) ellipse[x radius = 3, 
		y radius = 2];
		\node [at={(1.05,1.35)}] {$Q$};
		
		\draw (3,2.5) ellipse[x radius = 2.5, 
		y radius = 1.4];
		\node [at={(1.1,2.3)}] {$P$};
				
		\draw [] (6,1) ellipse[x radius = 1, 
		y radius = 1];
		\node [fill=white, at={(6,.4)}, rounded corners] {$R$};
		
		\draw [] (6.5,4) ellipse[x radius = 0.7, 
		y radius = 0.5];
		\node [fill=white, at={(6.5,4)}, rounded corners] {$S$};
		
	\end{tikzpicture}
\end{wrapfigure}

The OOD precise definition (\ie what is \textit{not} in-distribution) is elusive. If the underlying task and data distribution are non-overlapping ($S$), the question is settled. However, some OOD samples (such as adversarial examples) are very close to actual ID samples ($R$) such that, for example, in the absence of semantic changes, a human can predict the true label despite the adversarial noise. On the other end of the spectrum, there is no question that clean training samples ($P$) are ID. However, it is considered beneficial for classifiers to be robust against distribution shifts ($Q$) and generalize beyond their training sets. Most works in anomaly detection and/or robust classification are different in how they define what is an appropriate definition for $Q$, i.e., the distribution (or set) of samples that should be classified at test (inference) time. Importantly, these approaches also differ in how they define $Q$. \citeauthor{goldwasser2020beyond} directly work with samples $\tilde{x} \sim Q$ and present two algorithms given labels are present for $\tilde{x}$ or not.  If we let $S = \{ x | m(x) = \top\}$ and $\omega = 1/2$, our empirical loss function in \eqref{eq:loss} is similar to that of a PQ-learner. The key difference is that PQ-learners studied in~\cite{goldwasser2020beyond} do not have bounded type I error in their  selection step.

Note that for ease of presentation, in \eqref{eq:loss}, we defined $Q$ by the ground truth membership labels $z$. However, in our method in \S\ref{sec:method}, we do not require such explicit labels. Instead, the choice of distance metric, the k-nearest neighbor graph in each layer, the choice of standard hypothesis tests, and the aggregation method implicitly dictate our effective $Q$ distribution.

\section{Empirical Evaluation}

\paragraph{Gaussians}

\begin{table}
  \caption{\textbf{Experimental results on synthetic Gaussians with \pdknn and competing methods.} For in-distribution data (\texttt{gauss}), we \underline{align} the percentage of accepted samples to 0.965 and report the classification accuracy. For test-time data (\texttt{g\_1}, \texttt{g\_2}, \texttt{g\_3}), we only report the ratio of accepted samples. Up arrows ($\uparrow$) indicates that higher rates are better, down arrows ($\downarrow$) indicates that lower rates are better, right arrows ($\rightarrow$) indicates that a balanced rate (0.5) is best.}
  \label{tab:gaussian_main}
  \vspace{5pt}
  \centering
  \begin{tabular}{cccccc}
    \toprule
    & \multicolumn{2}{c}{\texttt{gauss}} & \multicolumn{1}{c}{\texttt{g\_1}} & \multicolumn{1}{c}{\texttt{g\_2}} & \multicolumn{1}{c}{\texttt{g\_3}}\\
    \cmidrule(r){2-3}
    \cmidrule(r){4-4}
    \cmidrule(r){5-5}
    \cmidrule(r){6-6}
    Model & Pass & Acc & Pass & Pass & Pass\\
    \midrule
    NN \citep{hendrycks2016baseline} & \underline{0.965} & \textbf{0.997} $\uparrow$ & 1.0 $\downarrow$ & 0.915 $\downarrow$ & 0.496 $\rightarrow$ \\
    DkNN \citep{dknn} & \underline{0.965} & 0.996 $\uparrow$ & 1.0 $\downarrow$ & 0.999 $\downarrow$ & 0.751 $\rightarrow$ \\
    ReBeL \citep{raghuram2020detecting} & \underline{0.965} & 0.991 $\uparrow$ & 0.975  $\downarrow$ & 0.871 $\downarrow$ & 0.667 $\rightarrow$  \\
    Trust \citep{jiang2018trust} & \underline{0.965} & N/A & 0.04 $\downarrow$ & 0.005 $\downarrow$ & 0.533 $\rightarrow$ \\
    \pdknn & \underline{0.965} & \textbf{0.997} $\uparrow$ & \textbf{0.0} $\downarrow$ & \textbf{0.0} $\downarrow$ & \textbf{0.498} $\rightarrow$ \\
    \bottomrule
  \end{tabular}
\end{table}

\begin{figure*}[t]
    \centering
    \begin{subfigure}[t]{0.31\textwidth}
        \centering
        \includegraphics[width=.75\linewidth]{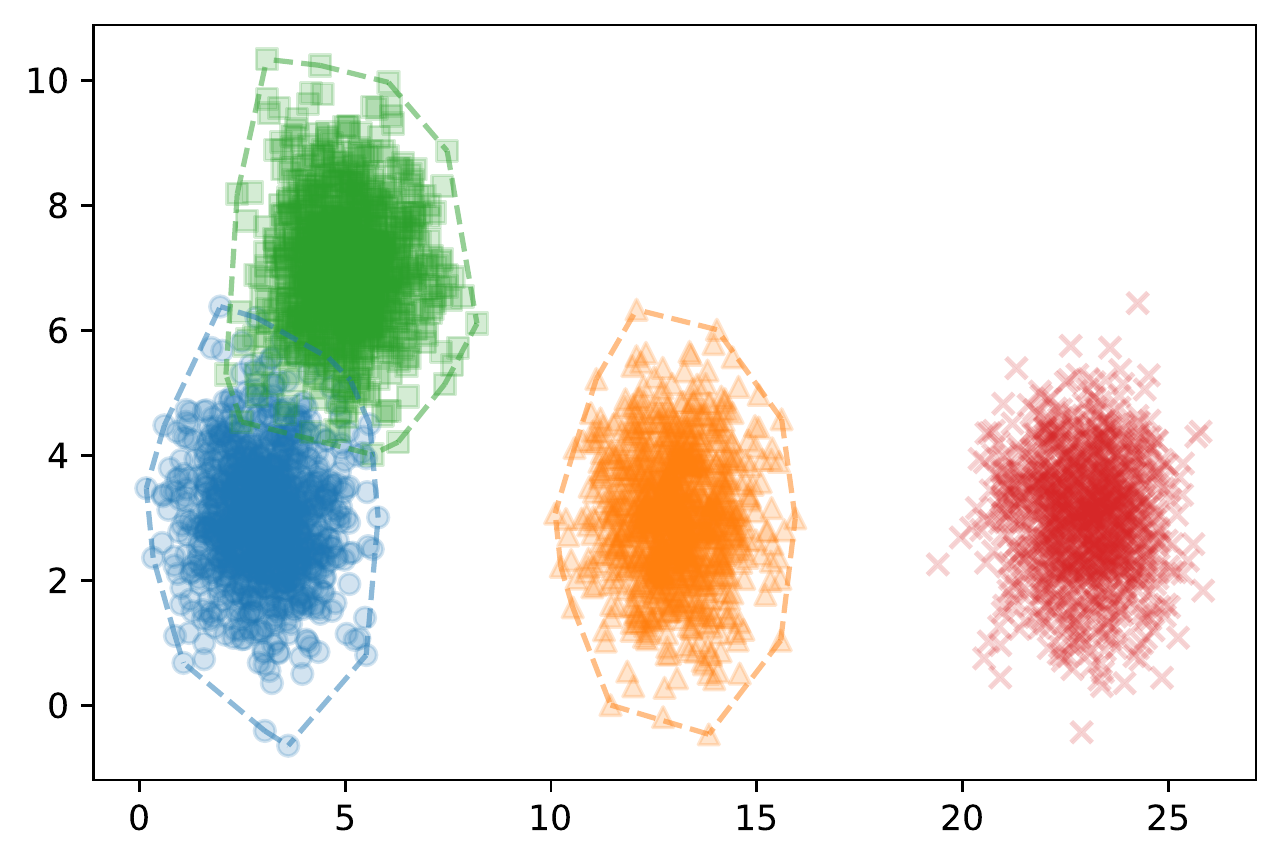}
    \end{subfigure}%
    ~
    \begin{subfigure}[t]{0.31\textwidth}
        \centering
        \includegraphics[width=.75\linewidth]{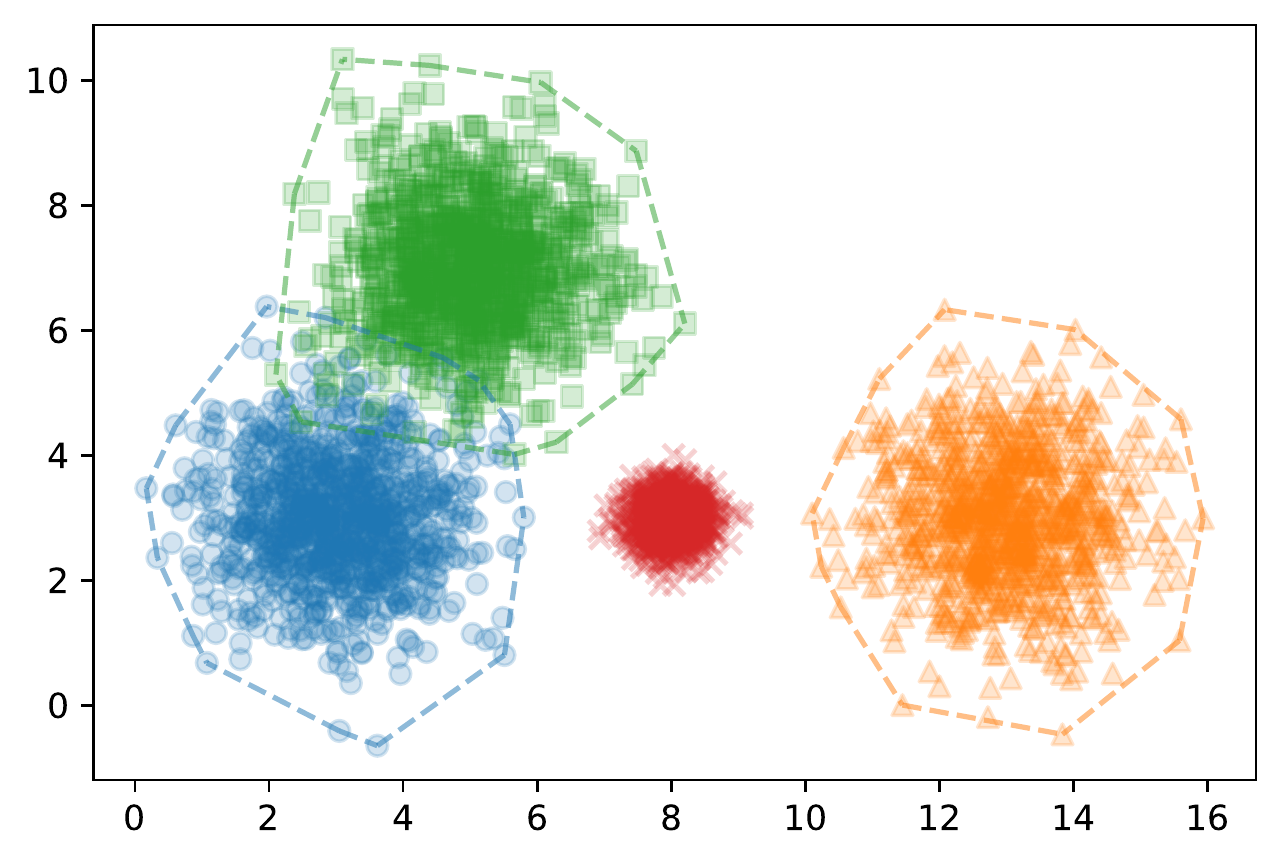}
    \end{subfigure}
    ~
    \begin{subfigure}[t]{0.31\textwidth}
        \centering
        \includegraphics[width=.75\linewidth]{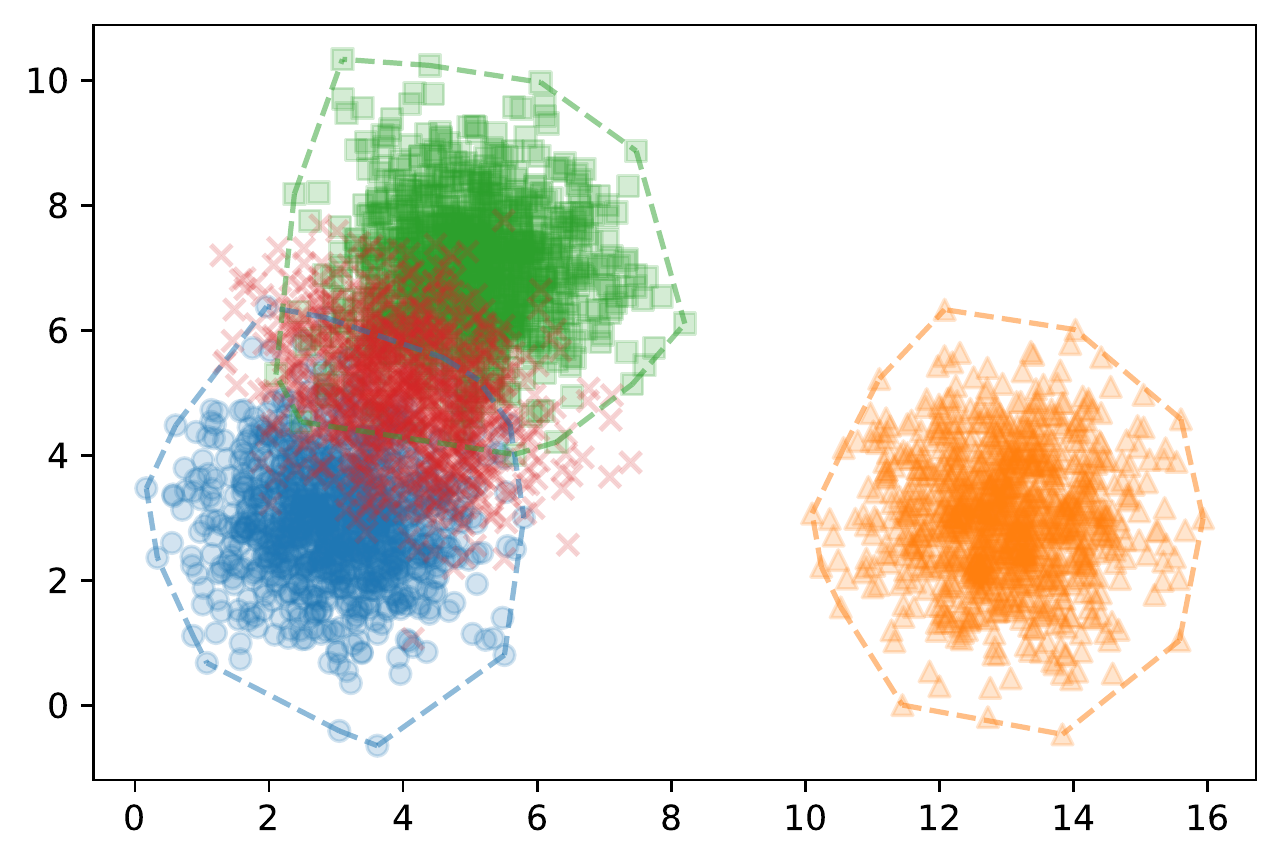}
    \end{subfigure}
    
    \begin{subfigure}[t]{0.31\textwidth}
        \centering
        \includegraphics[width=.75\linewidth]{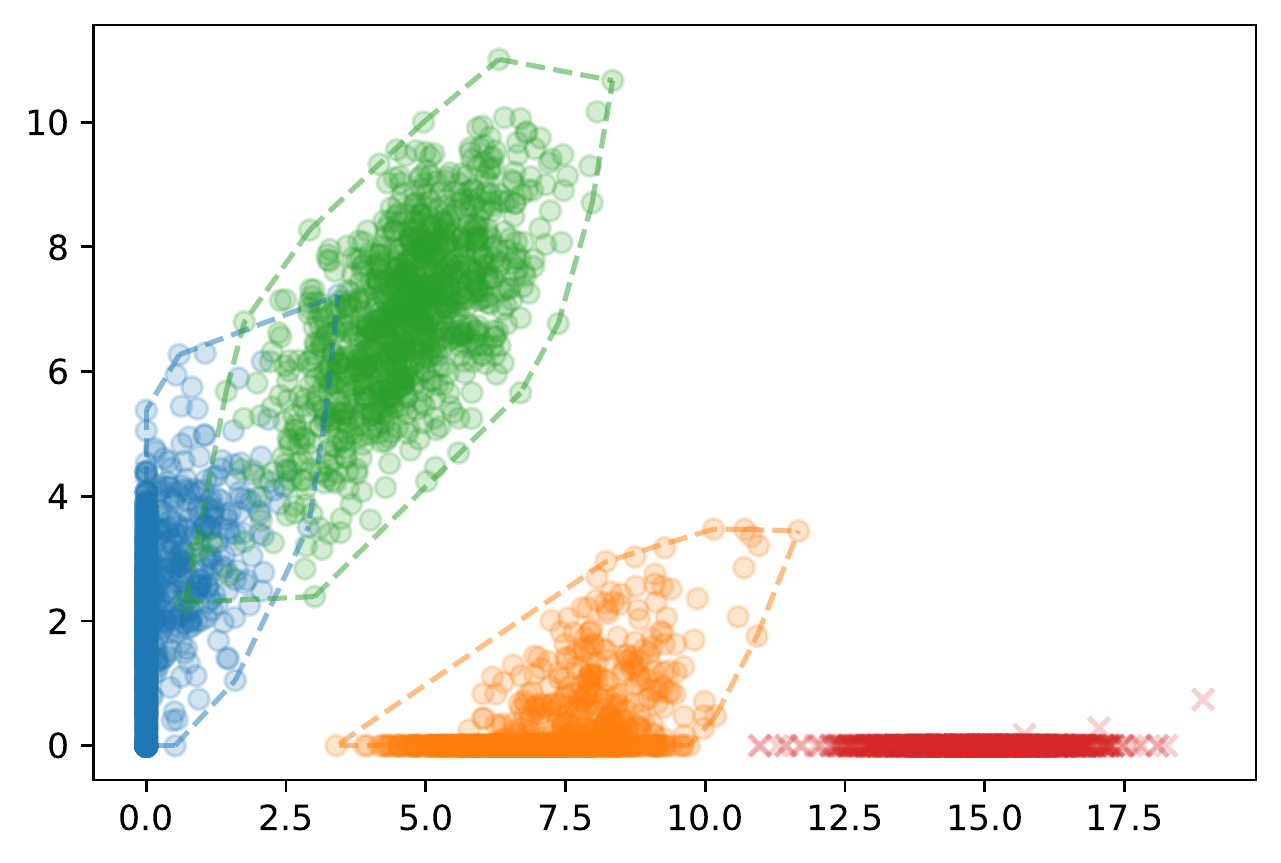}
    \end{subfigure}
    ~
    \begin{subfigure}[t]{0.31\textwidth}
        \centering
        \includegraphics[width=.75\linewidth]{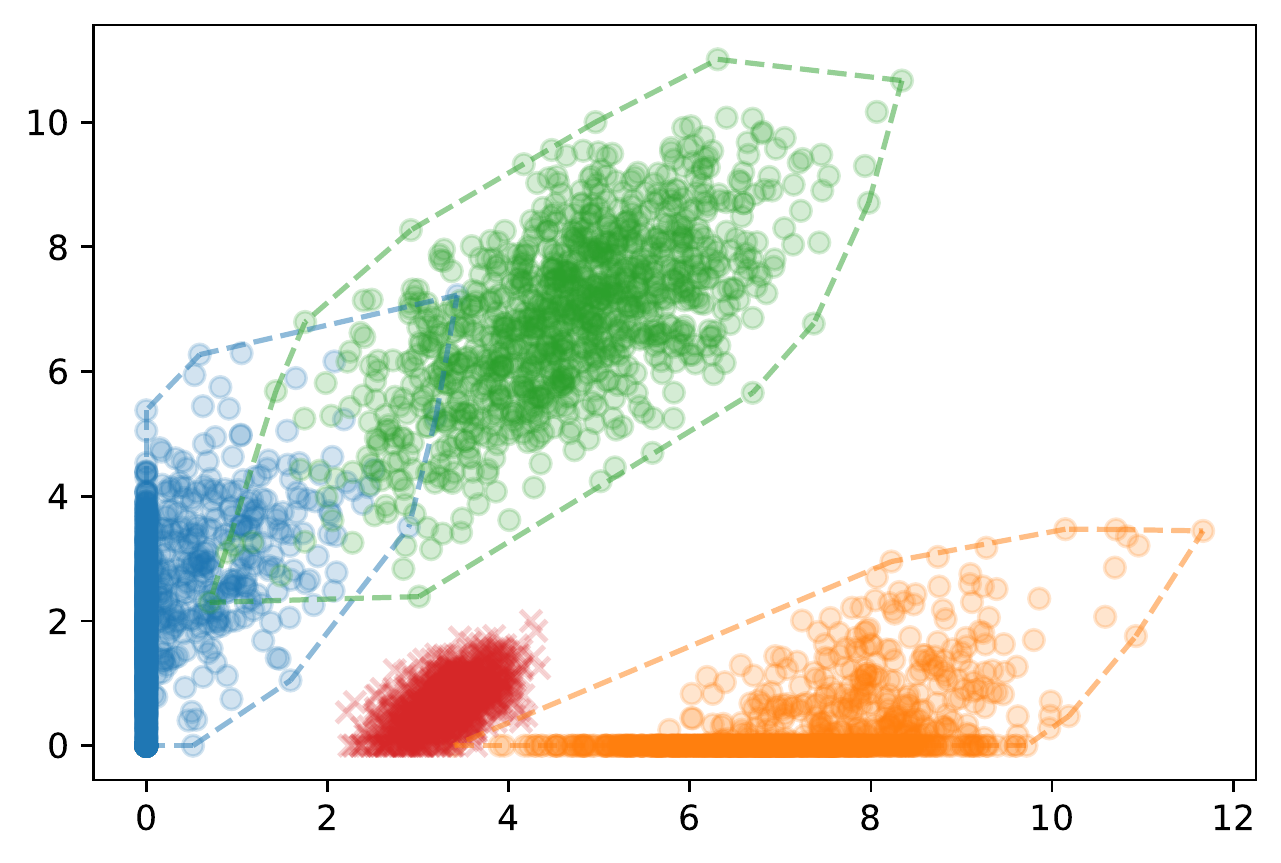}
    \end{subfigure}
    ~
    \begin{subfigure}[t]{0.31\textwidth}
        \centering
        \includegraphics[width=.75\linewidth]{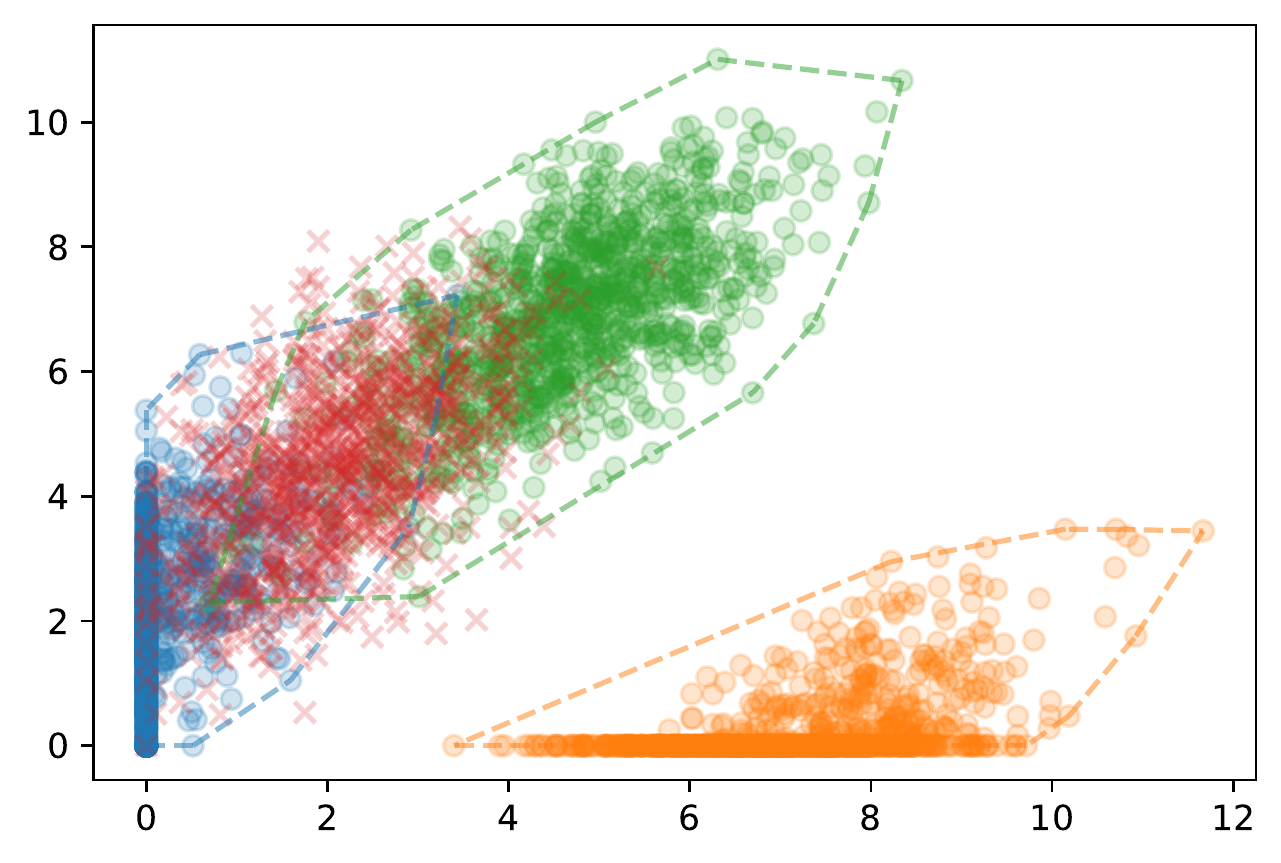}
    \end{subfigure}
    
    \begin{subfigure}[t]{0.31\textwidth}
        \centering
        \includegraphics[width=.75\linewidth]{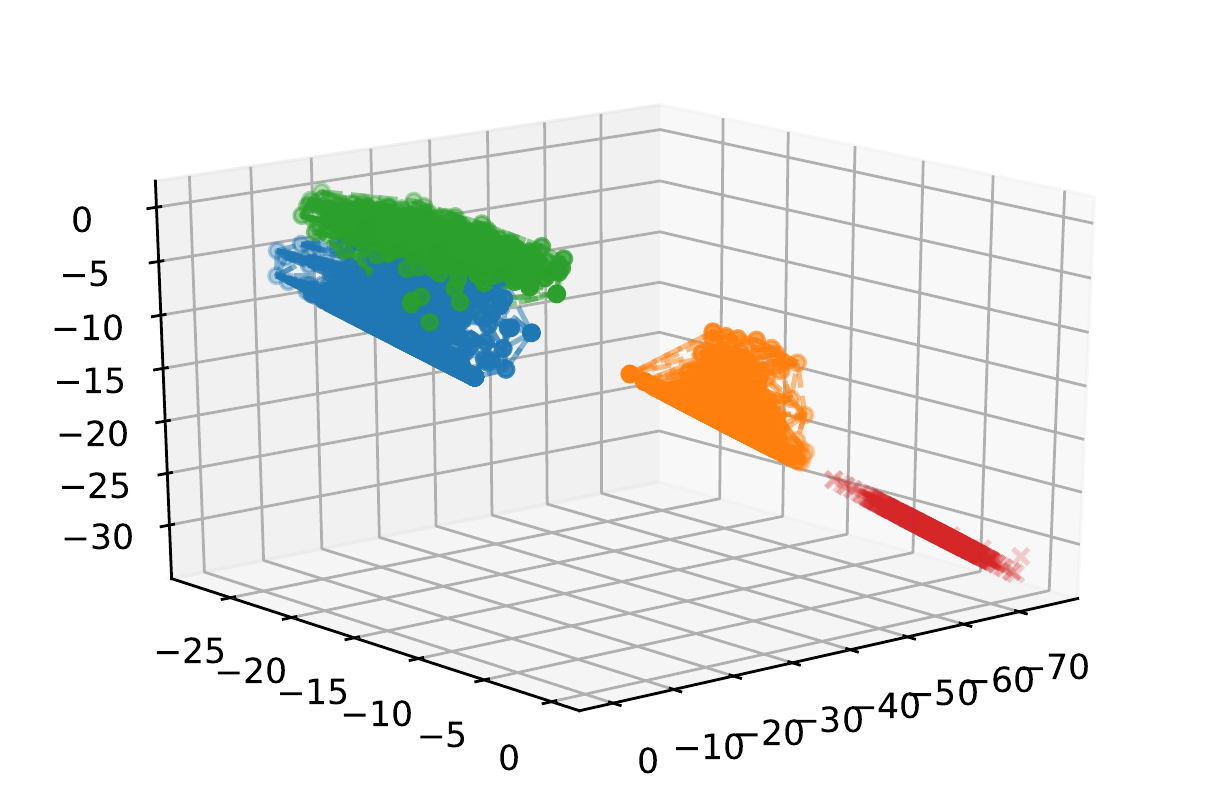}
        \caption{\texttt{g\_1}}
    \end{subfigure}
    ~
    \begin{subfigure}[t]{0.31\textwidth}
        \centering
        \includegraphics[width=.75\linewidth]{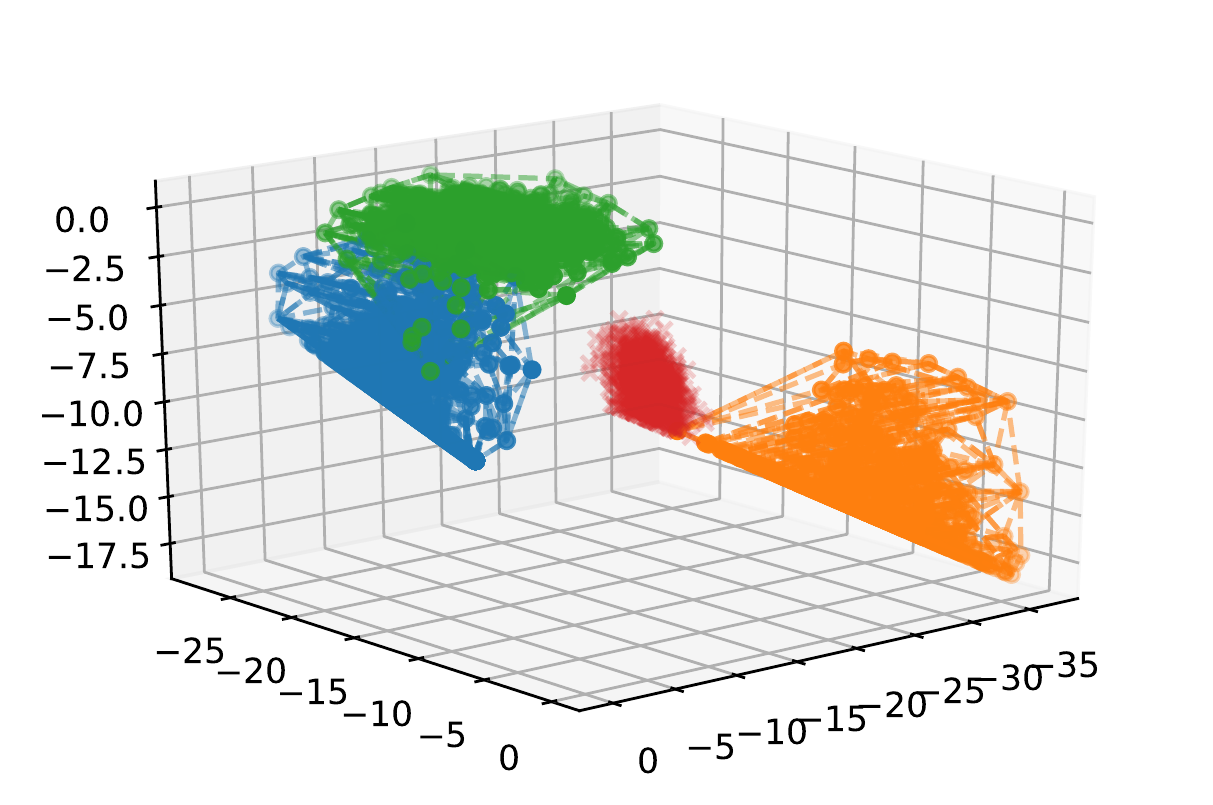}
        \caption{\texttt{g\_2}}
    \end{subfigure}
    ~
    \begin{subfigure}[t]{0.31\textwidth}
        \centering
        \includegraphics[width=.75\linewidth]{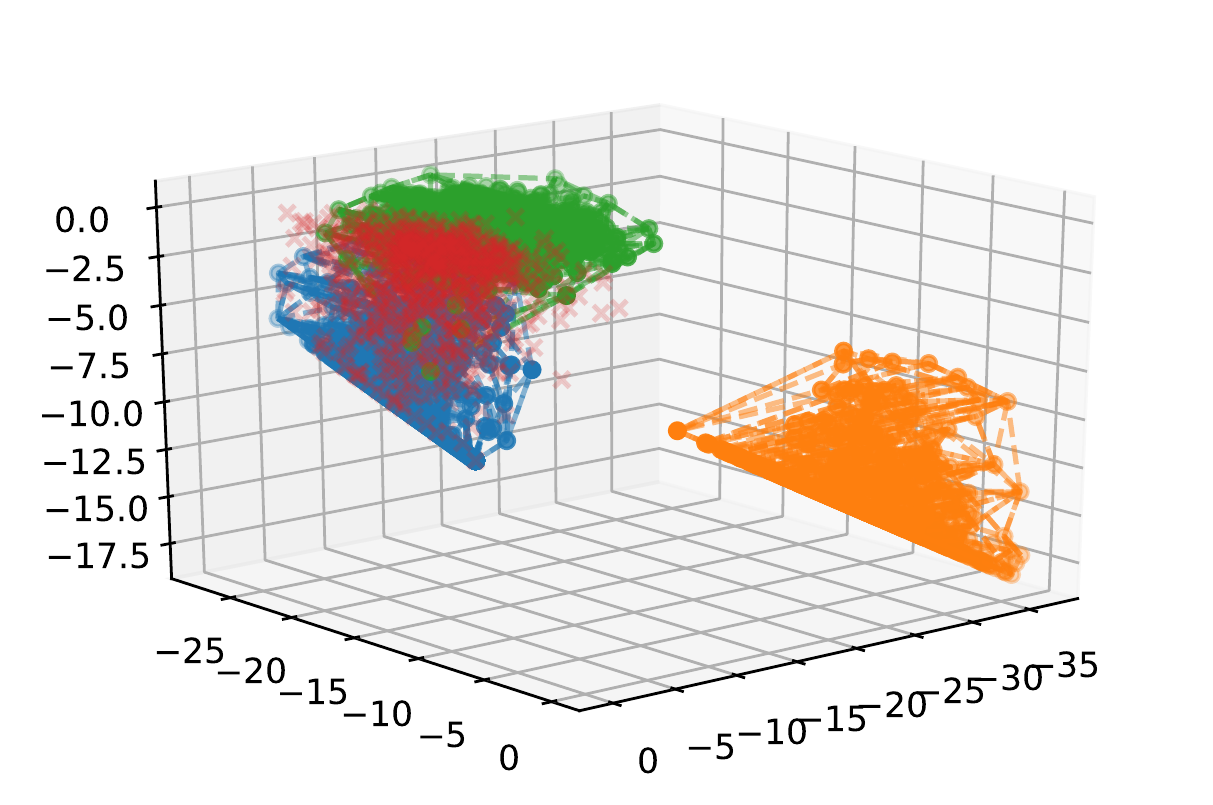}
        \caption{\texttt{g\_3}}
    \end{subfigure}
    \caption{\textbf{Neural network representations for our \texttt{gauss} datasets}. Circles \textcolor{blue}{$\circlet$} ($g_1$), triangles \textcolor{orange}{$\trianglepa$} ($g_2$), and squares \textcolor{green}{$\squad$} ($g_3$) show ID training data, while crosses \textcolor{red}{$\scross$} show (potentially OOD) test-time inputs. The first row depicts the input space for the three datasets, the second row the first hidden layer, and the third row the log-softmax output. Dashed lines are the convex hulls.}
    \label{fig:gauss_act}
\vskip -0.2in
\end{figure*}

\begin{table*}[!ht]
\caption{\textbf{Performance Comparison} between standard NN (Neural Network) architecture, \dknn and \pdknn. $\uparrow$ higher is better, $\downarrow$ lower is better, otherwise not applicable. Pairs of OOD data (OOD-1, OOD-2) for: MNIST (NotMNIST, FashionMNIST), FashionMNIST (NotMNIST, MNIST), SVHN (\cifarX, GTSRB), GTSRB (\cifarX, SVHN). Rotated samples are denoted as $\measuredangle$45. We report the rate of examples that passed the threshold (\textit{Pass}) and their accuracy (\textit{Acc}) (on the accepted examples). The results are \underline{aligned} according to \textit{Pass} on the clean samples (Attack None). Exact hyperparameter choices for all methods are documented in Section~\ref{sec:details_hyper}.}
\label{tab:compare}
\vskip -0.2in
\begin{center}
\begin{small}
\begin{sc}
 \begin{tabular}{l l  c c  c c  c c c c } 
 \toprule
  \multicolumn{2}{c}{} & \multicolumn{2}{c}{\mnist} & \multicolumn{2}{c}{\fashion} & \multicolumn{2}{c}{SVHN} & \multicolumn{2}{c}{GTSRB} \\
 \cmidrule(r){3-4}
 \cmidrule(r){5-6}
 \cmidrule(r){7-8}
 \cmidrule(r){9-10}
 \textbf{Attack} & \textbf{Model} & \textbf{Pass} & \textbf{Acc} & \textbf{Pass} & \textbf{Acc} & \textbf{Pass} & \textbf{Acc} & \textbf{Pass} & \textbf{Acc} \\ [0.5ex] 
 \midrule
 \multirow{3}{*}{None} & NN & \underline{.956} $\uparrow$ & \textbf{.998} $\uparrow$ & \underline{.908} $\uparrow$ & .945 $\uparrow$ & \underline{.900}  $\uparrow$ & .949 $\uparrow$ & \underline{.900} $\uparrow$ & .928 $\uparrow$ \\
 & \dknn & \underline{.956} $\uparrow$ & .996 $\uparrow$ & \textbf{\underline{.909}} $\uparrow$ & .936 $\uparrow$ & \underline{.892} $\uparrow$ & \textbf{.951} $\uparrow$ & \underline{.900} $\uparrow$ & \textbf{.979} $\uparrow$ \\  
& \pdknn & \underline{.956} $\uparrow$ & \textbf{.998} $\uparrow$ &  \underline{.908} $\uparrow$ & \textbf{.946} $\uparrow$ & \underline{.900} $\uparrow$ & .944 $\uparrow$ & \underline{.888} $\uparrow$ & .970 $\uparrow$ \\   
\midrule
\multirow{3}{*}{OOD-1} & NN & .200 $\downarrow$ & .104 & .774 $\downarrow$ & .150 & .521 $\downarrow$ & .093 & \textbf{.169} $\downarrow$ & .004 \\
 & \dknn & .235 $\downarrow$ & .163 & .414 $\downarrow$ & .283 & \textbf{.322} $\downarrow$ & .090 & .281 $\downarrow$ & .001 \\
 & \pdknn & \textbf{.182} $\downarrow$  & .154 & \textbf{.405} $\downarrow$ & .257 & .375 $\downarrow$ & .094 & .269 $\downarrow$ & .004\\
 \midrule
  \multirow{3}{*}{OOD-2} & NN & \textbf{.068} $\downarrow$ & .047 & .748 $\downarrow$ & .130 & .545 $\downarrow$ & .034 & \textbf{.214} $\downarrow$ & .003 \\
 & \dknn & .300 $\downarrow$ & .034 &  .441 $\downarrow$ & .148 & \textbf{.330} $\downarrow$ & .017 & .309 $\downarrow$  & .003\\
& \pdknn & .170 $\downarrow$ & .042 &  \textbf{.161} $\downarrow$ & .263 & .362 $\downarrow$ & .020 & .287 $\downarrow$ & .003 \\
 \midrule
\multirow{3}{*}{Feat} & NN & \textbf{.0} $\downarrow$ & .0 & .471 $\downarrow$ & .0  & .702 $\downarrow$ & .0 & \textbf{.362} $\downarrow$ & .0 \\
 & \dknn & .442 $\downarrow$ & .732 & .716 $\downarrow$ & .523 & .686 $\downarrow$ & .373 & .516 $\downarrow$ & .102 \\
 & \pdknn & .161 $\downarrow$ & .133 & \textbf{.389} $\downarrow$ & .146 & \textbf{.571} $\downarrow$ & .303 & .452 $\downarrow$ & .142 \\
\midrule
\multirow{3}{*}{FGSM} & NN & .680 $\downarrow$ & .022 & .933 $\downarrow$ & .095 & .828 $\downarrow$ & .042 & .489 $\downarrow$ & .094\\
 & \dknn & .373 $\downarrow$ & .149 & .672 $\downarrow$ & .398 & \textbf{.485} $\downarrow$ & .046 & .417 $\downarrow$ & .199\\
 & \pdknn & \textbf{.239} $\downarrow$ & .102 & \textbf{.589} $\downarrow$ & .170 & .502 $\downarrow$ & .039 & \textbf{.379} $\downarrow$ & .191 \\
 \midrule
 \multirow{3}{*}{PGD} & NN & .999  $\downarrow$ & .010 & 1.00 $\downarrow$ & .061 &1.00 $\downarrow$ & .030 & 1.00 $\downarrow$ & .012 \\
 & \dknn & .646  $\downarrow$ & .014 & .693 $\downarrow$ & .164 & .946 $\downarrow$ & .030 & .850 $\downarrow$ & .018 \\
 & \pdknn & \textbf{.518} $\downarrow$ & .016 & \textbf{.575} $\downarrow$ & .098 & \textbf{.929} $\downarrow$ & .031 & \textbf{.826} $\downarrow$ & .016 \\
 \midrule
  \multirow{3}{*}{$\measuredangle$45} & NN & \textbf{.399} $\downarrow$ & .816 & .675 $\downarrow$ & .140 & .535 $\downarrow$ & .351 & .556 $\downarrow$ & .180 \\
 & \dknn & .439 $\downarrow$ & .788 & \textbf{.251} $\downarrow$ & .353 & \textbf{.165} $\downarrow$ & .362 & .572 $\downarrow$ & .197 \\
& \pdknn & .440 $\downarrow$ & .846  & .307 $\downarrow$ & .243 & .273 $\downarrow$ & .359 & \textbf{.502} $\downarrow$ & .212  \\
 \bottomrule
\end{tabular}
\end{sc}
\end{small}
\end{center}
\vskip -0.1in
\end{table*}

We first want to determine how well \pdknn handles various types of carefully crafted (potentially OOD) test-time inputs and how specifically OOD decisions are being made by \pdknn. To do so, we evaluate our approach on a synthetic dataset consisting of three two-dimensional Gaussian blobs: $g_1 \sim \mathcal{N}(\begin{bmatrix}3 & 3\end{bmatrix}^T,\bm{I})$, $g_2 \sim \mathcal{N}(\begin{bmatrix}13 & 3\end{bmatrix}^T,\bm{I})$, $g_3 \sim \mathcal{N}(\begin{bmatrix}5 & 7\end{bmatrix}^T,\bm{I})$. We train a neural network classifier consisting of a single fully-connected two-dimensional latent layer and a fully-connected three-dimensional log-softmax output layer using 1000 samples from each Gaussian and achieve 98.80\% accuracy on this task. This network can be used as a stand-alone classifier or can be augmented using \pdknn's inference procedure.

To test the rejection rate and accuracy of \pdknn we generate four test sets: (i) \texttt{gauss} consisting of in-distribution data points from Gaussians $g_1$, $g_2$, and $g_3$, a clear ID dataset; (ii) \texttt{g\_1} consisting of data points sampled from $\mathcal{N}(\begin{bmatrix}23 & 3\end{bmatrix}^T,\bm{I})$, a clear OOD dataset; (iii) \texttt{g\_2} consisting of data points sampled from $\mathcal{N}(\begin{bmatrix}8 & 3\end{bmatrix}^T,0.1 \bm{I})$, another OOD dataset; and (iv) \texttt{g\_3} consisting of data points sampled from $\mathcal{N}(\begin{bmatrix}4 & 5\end{bmatrix}^T,\begin{bmatrix}1 & -0.2\\-0.2 & 1\end{bmatrix})$, a dataset that is designed to spread its density equally between the intersection of $g_1$ and $g_2$ (a region of high classification uncertainty) and the individual clusters $g_1$ and $g_2$ (regions of low classification uncertainty). All datasets, including their neural network activations, are shown in Figure \ref{fig:gauss_act}.

We report our results in detail in Table \ref{tab:gaussian_main}. In addition to thresholding the underlying neural network \citep{hendrycks2016baseline} and applying \dknn on top of the underlying neural network, we also compare against two recent related OOD detection techniques: \textsc{ReBeL} \citep{raghuram2020detecting} and \textsc{Trust} \citep{jiang2018trust}. We clearly see that competing approaches are outperformed by \pdknn on all datasets.

In addition to the acceptance rate of and the accuracy on samples, we also analyze the rejection reason for rejected data points for \pdknn.  
For \texttt{gauss} and \texttt{g\_3}, i.e. datasets that completely lie within the support of the trained distribution, all rejections happen because of inconclusive \pvalues. On the other hand, all rejections are determined by an exceeded distance to the convex hull of the nearest cluster for \texttt{g\_1}. \texttt{g\_2}, a dataset that is positioned exactly between the three in-distribution Gaussians, strikes a middle-ground as 53\% of rejections happen because of inconclusive \pvalues (data points close to the mode of the Gaussian) while 47\% of rejections can be attributed to the convex hull criterion (data points in the tail of the Gaussian).

\paragraph{Evaluation on Standard Datasets}
\label{sec:experiments}

Our method \pdknn builds upon \dknn. Thus, we evaluate \pdknn against \dknn as well as standard neural networks (NN) in Table~\ref{tab:compare}. We use \mnist, \fashion, SVHN, and GTSRB datasets as clean samples. We test the methods on  out-of-distribution (OOD), rotated samples, and adversarial examples found with a feature-based adaptive attack~\citep{feature2016attack} denoted as \textit{Feat} (details below), PGD~\citep{madry2018towards}, and FGSM~\citep{goodfellow2014explaining}, with max distortion $\varepsilon=0.3$ on MNIST and FashionMNIST, and $\varepsilon=0.03$ on SVHN and GTSRB.

\begin{wrapfigure}{r}{0.25\textwidth} 
	\vspace{-10pt}	
	\hspace{32pt}
	\scalebox{0.15}{
		\includegraphics[]{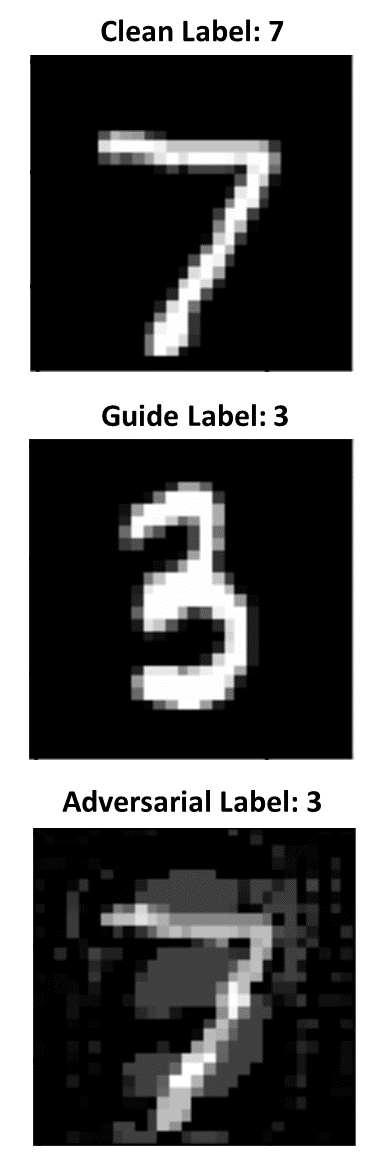}
	}
\caption{\textbf{Feature attack} with digit 3 for the digit 7 against \pdknn.\label{fig:feature-example}
}
\end{wrapfigure}

The aggregation of \pvalues is conservative and in practice requires us to weight the values from each layer differently. \citet{frosst19a} showed that there is a high entanglement in all layers apart from the last one. We use this result as our guidance in the selection of weights per layer. We weight the \pvalues based on the layers they come from and average them to reduce their magnitude.  

The feature attack is a natural choice for \dknn and \pdknn: it optimizes for a minimal perturbation that moves an internal representation of a clean sample close to an adversarial representation (from another example called the \textit{guide}). 
Feature adversarial examples against \dknn and \pdknn usually draw the adversarial digit in the background, which is clearly visible in most cases~(see an example in Figure~\ref{fig:feature-example}). This confirms previous findings~\citep{sitawarin2019robustness,sitawarin2020minimum}.
The feature attack is not optimal for vanilla DNNs, where  adversarial examples optimizing over the model's output work better (e.g., FGSM, PGD, etc.). For both, \mnist and GTSRB we observe that NN rejects most of the feature adversaries by significantly lowering its confidence.

 \begin{wrapfigure}{R}{0.5\textwidth}
 \hspace{15pt}
 \begin{minipage}{0.45\textwidth}
     \vspace{0pt}
 	\input{contents/pdknn_rebel_alg}
     \vspace{-0.2in}
 \end{minipage}
 \end{wrapfigure}

\paragraph{Comparison with \rebel and Trust} We also compare \pdknn against other unsupervised methods such as ReBeL \citep{raghuram2020detecting} and Trust \citep{jiang2018trust}. 
During said comparison, we adhere to their choice of basic parameters, such as the number of nearest neighbors $k$, which is given by a heuristic $k=\lceil n^{0.4} \rceil$, model architecture, and the tuning of hyper-parameters on the training set, where we adjust the \pvalue for \pdknn. This allows for a fair comparison between the methods.

\begin{figure}
    \centering
    \begin{subfigure}[b]{0.7\textwidth}
        \centering
        \includegraphics[width=\textwidth]{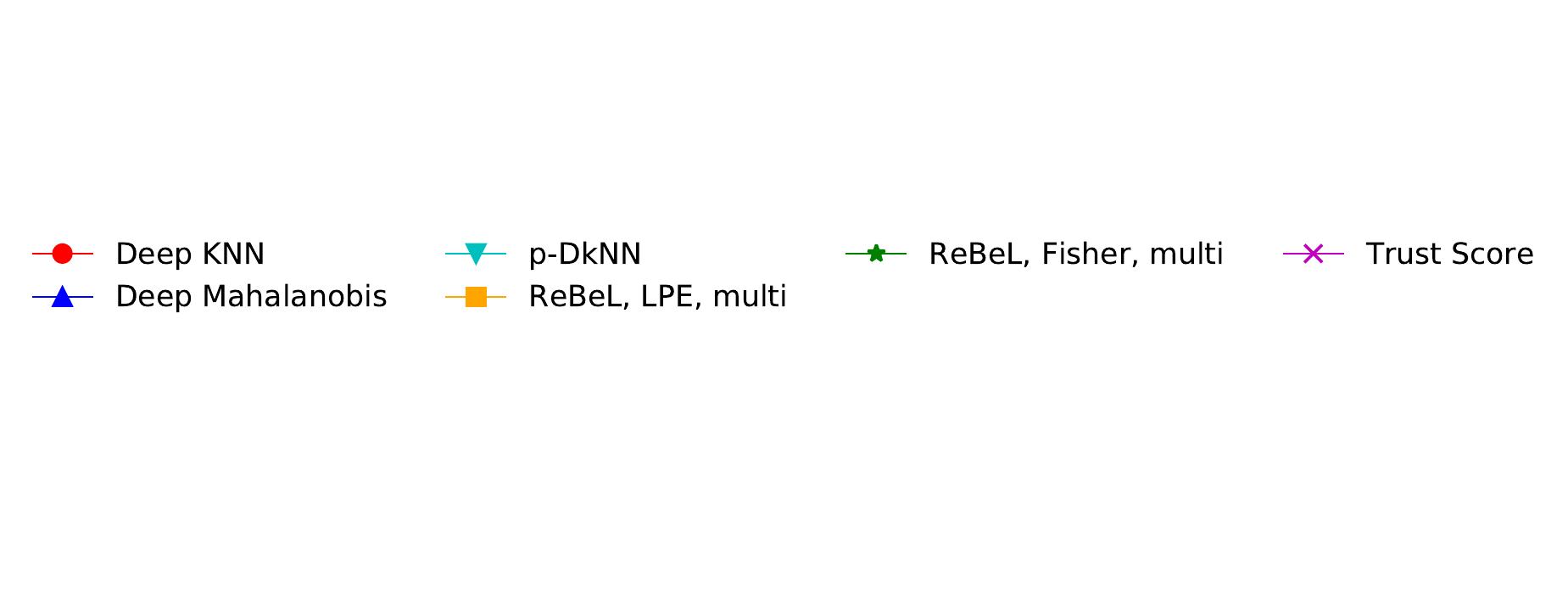}
    \end{subfigure}
    
    \vspace*{-4em}
    
    \begin{subfigure}[b]{0.32\textwidth}
        \centering
        \includegraphics[width=\textwidth]{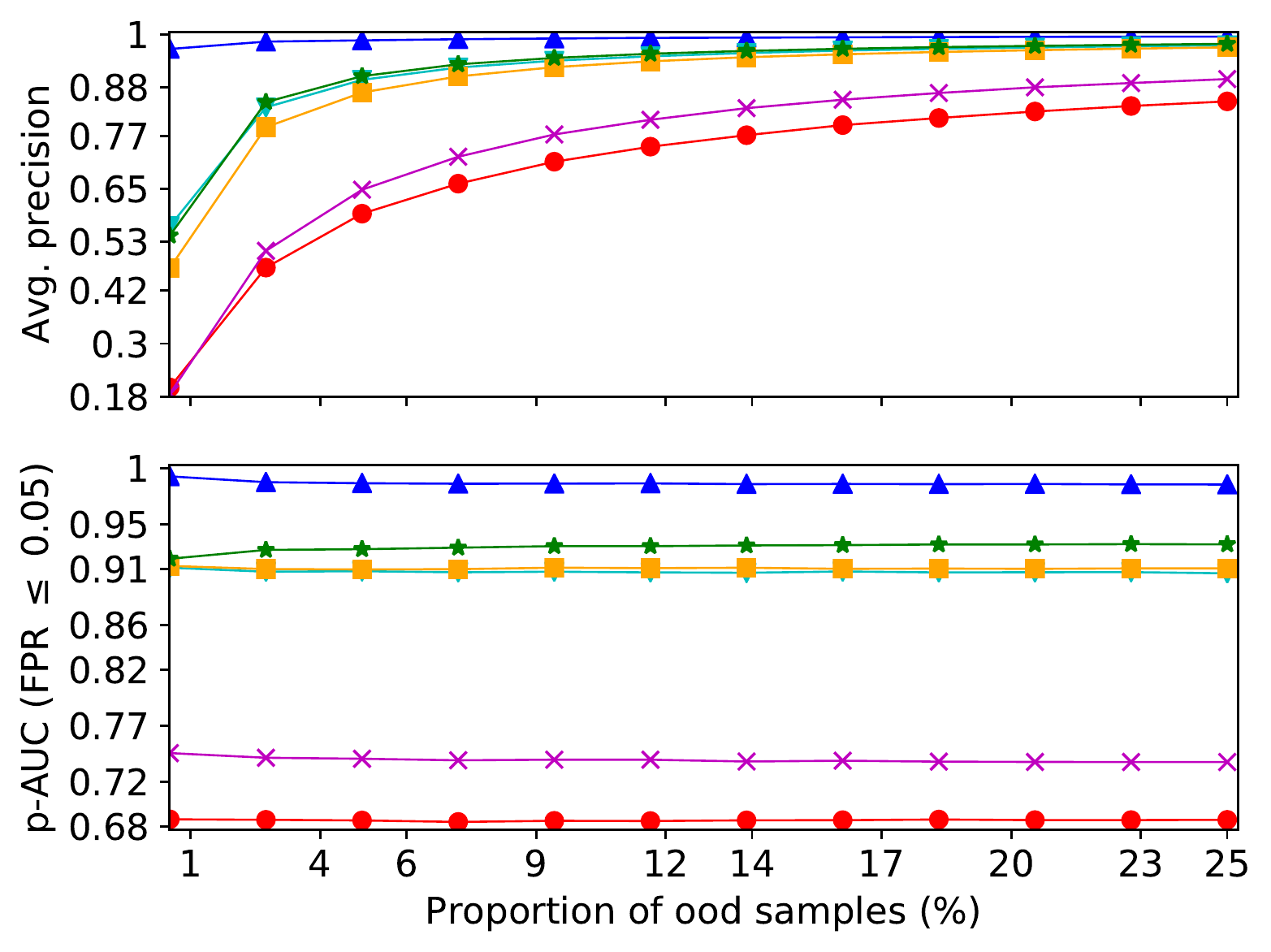}
        \caption{\mnist model vs notMNIST}
        \label{fig:mnist_ood_comparison}
    \end{subfigure}
    ~
    \begin{subfigure}[b]{0.32\textwidth}
        \centering
        \includegraphics[width=\textwidth]{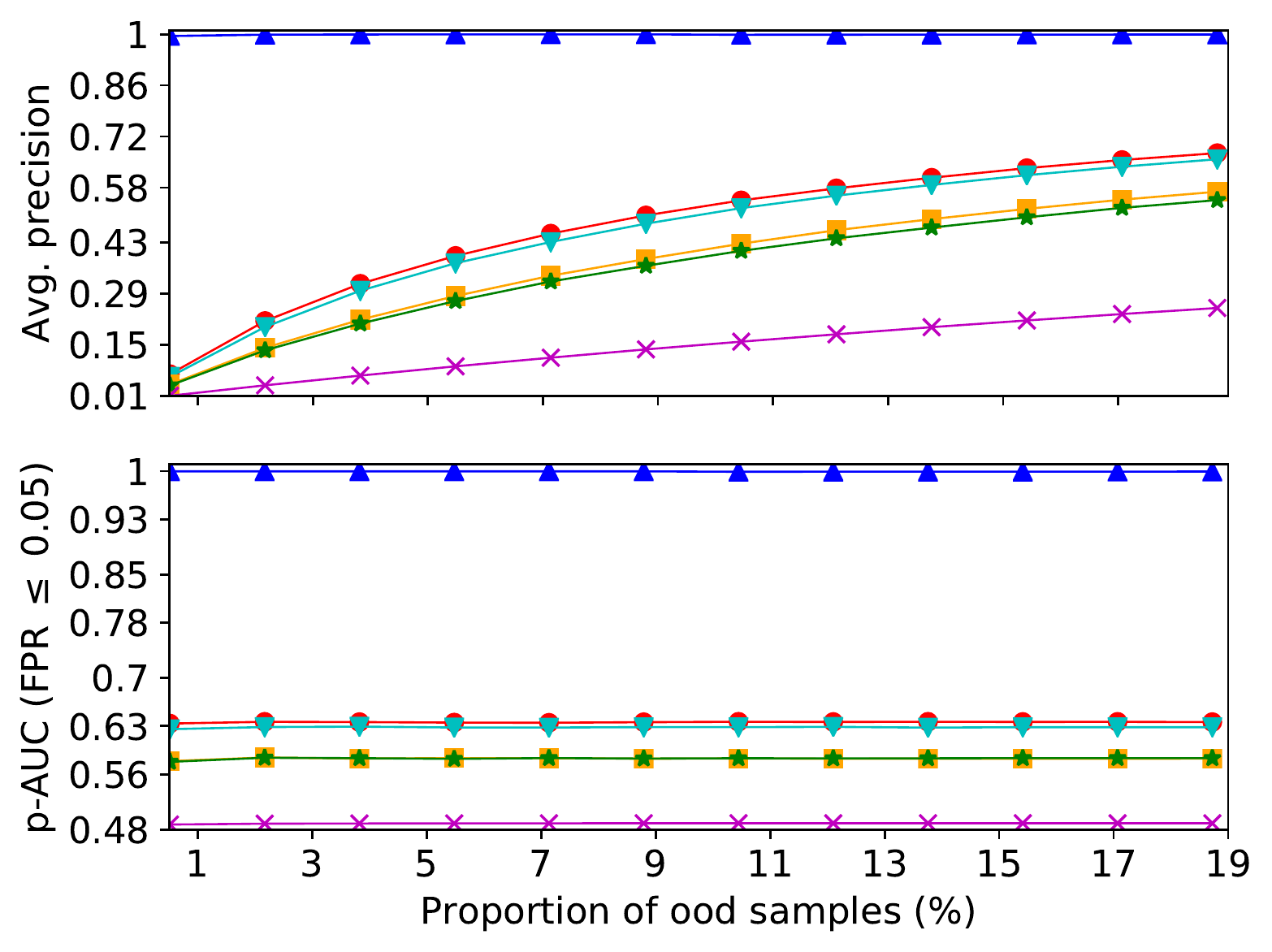}
        \caption{SVHN model vs \cifarX}
        \label{fig:svhn_ood_comparison}
    \end{subfigure}
    ~
    \begin{subfigure}[b]{0.32\textwidth}
        \centering
        \includegraphics[width=\textwidth]{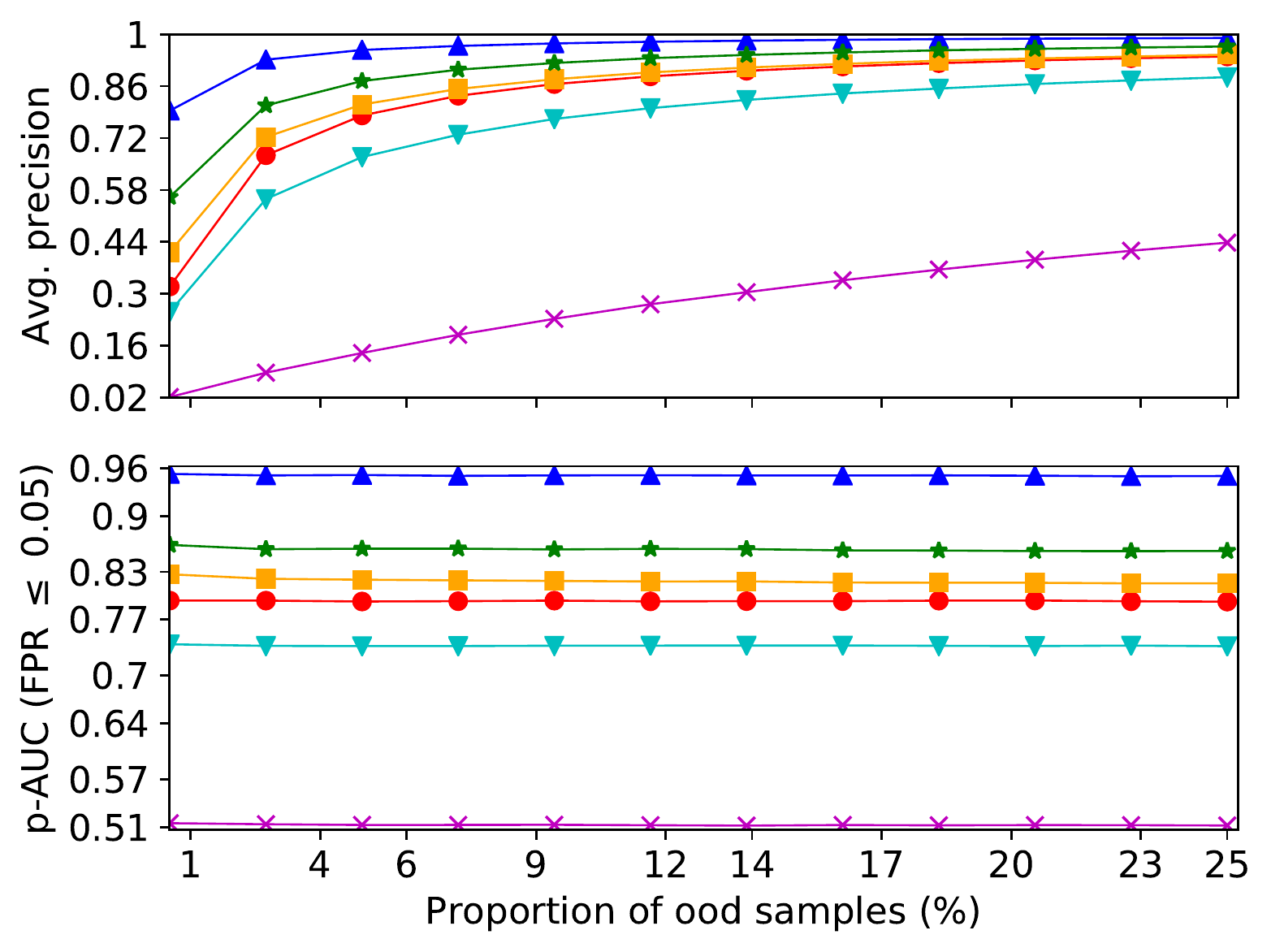}
        \caption{\cifarX model vs SVHN}
        \label{fig:cifar10_ood_comparison}
        \label{fig:ood-comparison}
    \end{subfigure}
    \caption{OOD detection performance comparison between \pdknn and other methods.}
\end{figure}

\begin{figure}
    \centering
    \begin{subfigure}[b]{0.7\textwidth}
        \centering
        \includegraphics[width=\textwidth]{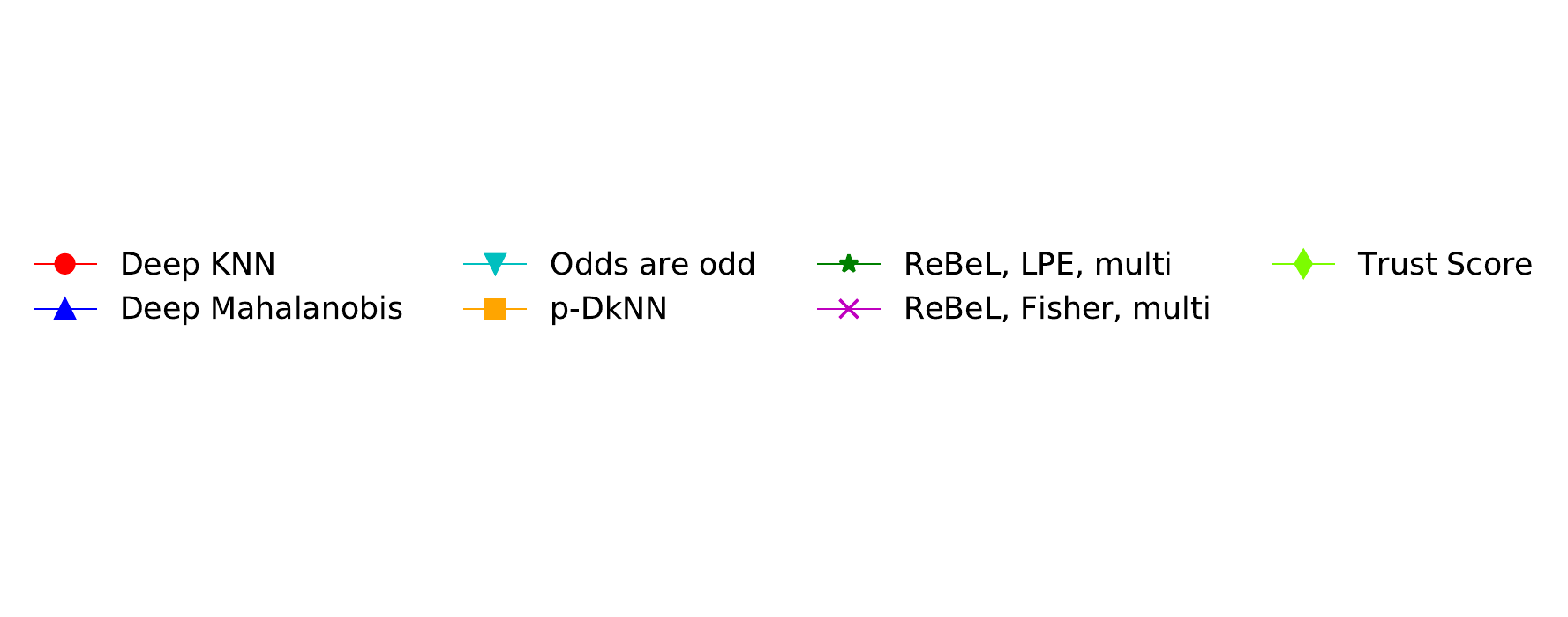}
    \end{subfigure}
    
    \vspace*{-4em}
    
    \begin{subfigure}[b]{0.31\textwidth}
        \centering
        \includegraphics[width=\textwidth]{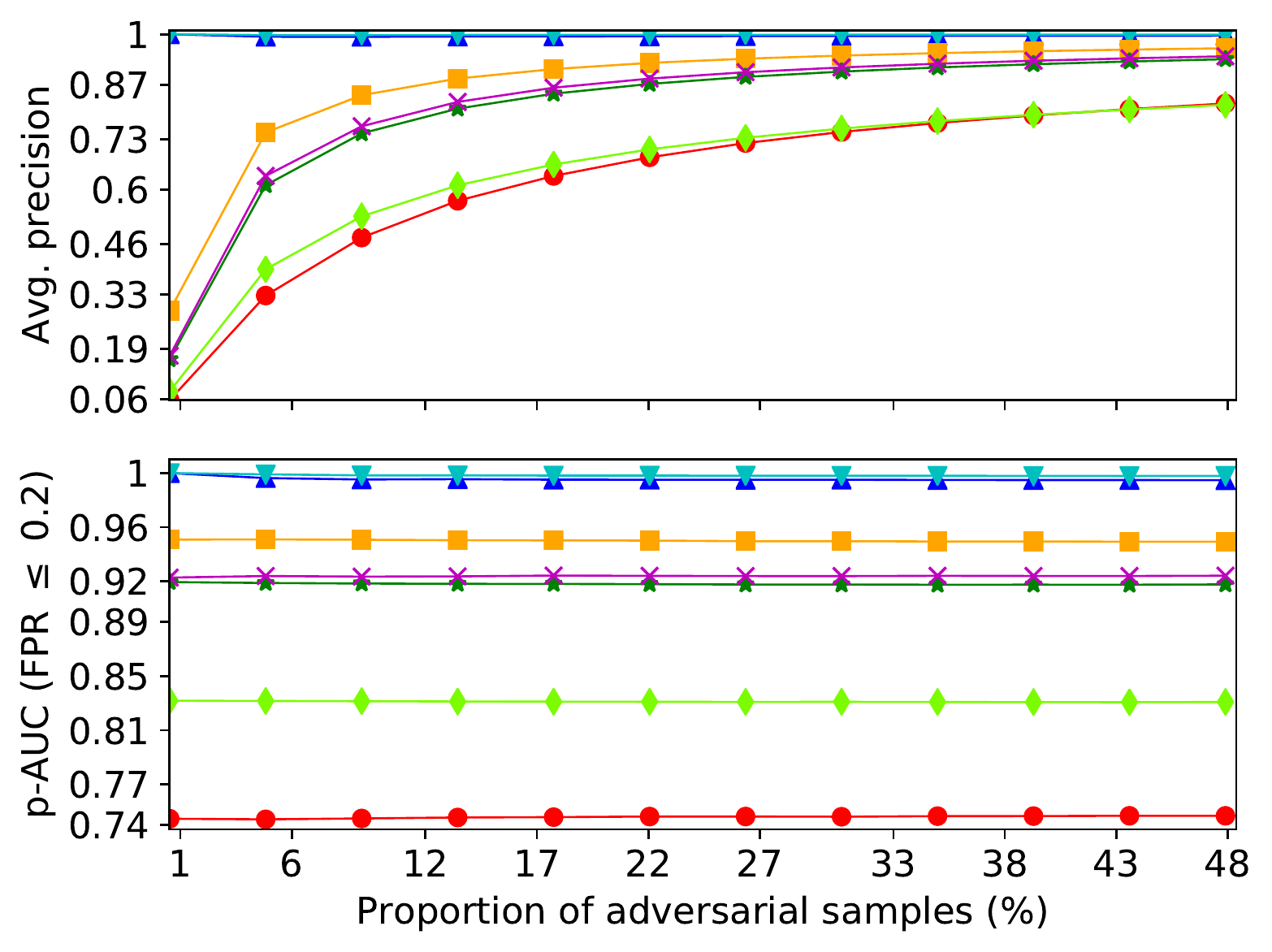}
        \caption{\mnist}
        \label{fig:mnist_adv_comparison}
    \end{subfigure}
    ~
    \begin{subfigure}[b]{0.31\textwidth}
        \centering
        \includegraphics[width=\textwidth]{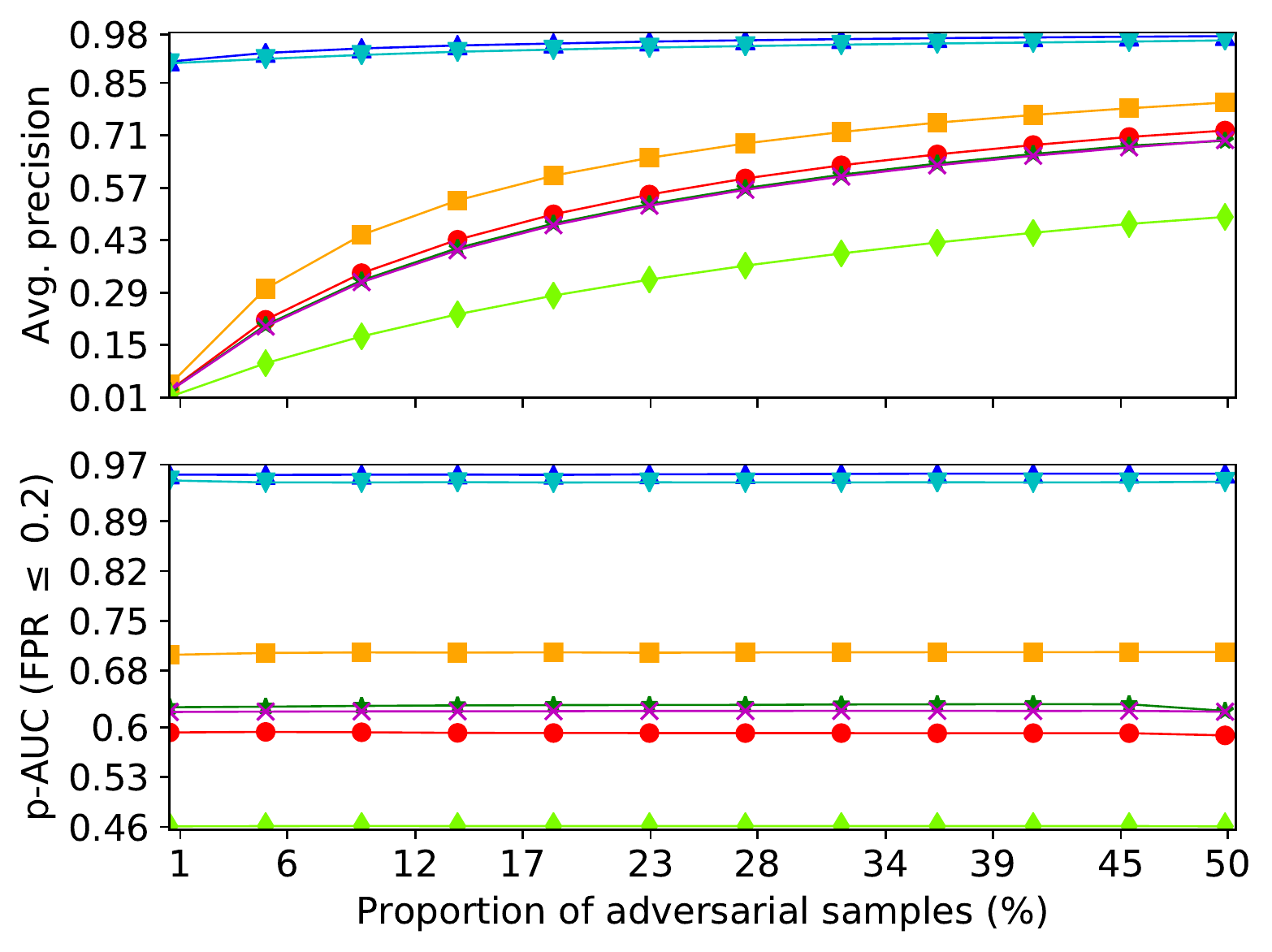}
        \caption{SVHN}
        \label{fig:svhn_adv_comparison}
    \end{subfigure}
    ~
    \begin{subfigure}[b]{0.31\textwidth}
        \centering
        \includegraphics[width=\textwidth]{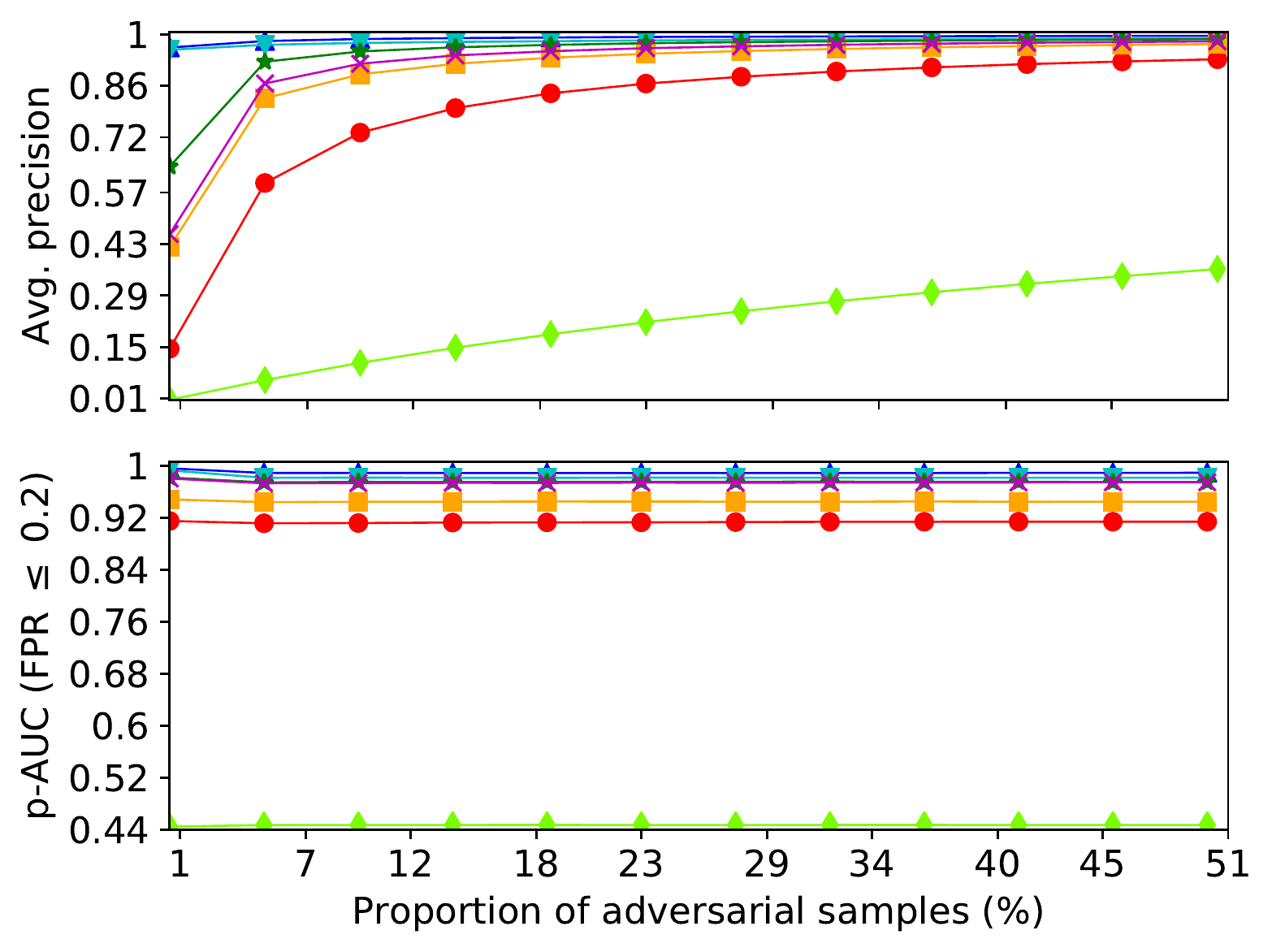}
        \caption{\cifarX}
        \label{fig:cifar10_adv_comparison}
    \end{subfigure}
    \label{fig:adv-comparison}
    \caption{PGD adversarial sample detection performance comparison between \pdknn and other methods.}
\end{figure}

For this comparison, we use all the components from our main method apart from the convex hulls and we find neighbors on the training set instead of the validation set. We present the adjusted method as Algorithm~\ref{alg:pdknn-rebel}. For the correction of \pvalues, we use the two stage fdr correction (non-negative), denoted as \textit{fdr\_tsbh}. We tune the weights per layer that are used for a weighted mean of the \pvalues and effect sizes. We find that for smaller networks, for instance, architectures used for MNIST or SVHN datasets with 2, 3 or 4 convolutional layers followed by 1 or 2 fully connected layers, it is beneficial to put more weight on the last layer. For larger networks, such as ResNet-34 used for \cifarX, the larger weights should be assigned to final layers (not only the last one).

First, we analyze the performance of \pdknn against other methods for OOD data. Our experiments in Figure~\ref{fig:mnist_ood_comparison} on MNIST show that \pdknn is on-par with ReBeL and outperforms other unsupervised methods (e.g., Trust Score). The Deep Mahalanobis~\citep{mahalanobis2018} method is supervised so it clearly performs better than the unsupervised methods. Then, for the SVHN dataset in Figure~\ref{fig:svhn_ood_comparison}, \pdknn is on par with \dknn and both methods are clearly better than other unsupervised ones. Finally, for \cifarX in Figure~\ref{fig:cifar10_ood_comparison}, \pdknn is slightly worse than ReBeL and DkNN but still performs much better than the Trust Score.

Second, for the adversarial examples found using the PGD attack, \pdknn is the best performing method for both MNIST (Figure~\ref{fig:mnist_adv_comparison}) and SVHN (Figure~\ref{fig:svhn_adv_comparison}) datasets. For the \cifarX dataset (Figure~\ref{fig:cifar10_adv_comparison}), our method is on par with ReBeL and outperforms other unsupervised methods. The Deep Mahalanobis and Odds are Odd methods are both supervised and clearly outperform the unsupservised methods. The Odds are Odd method targets specifically the adversarial examples, hence it is not applied to the OOD detection.

Overall, we find that \pdknn provides the best performance in most cases.
\section{Limitations}
\label{sec:limitations}
Due to our bounds on type I error, \pdknn is a conservative approach. The tight control on type I error is the bedrock of the hypothesis testing framework. Consequently, most techniques and considerations we delineated in Section~\ref{sec:method}---such as employing ANOVA tests, p-value corrections, and the use of a separate validation set to estimate representational distances---are meant to keep the type I error in check. However, as we demonstrated pictorially in Figure~\ref{fig:np-vs-normal} (d), a strict control on type I error would necessarily mean an increase in type II error (and by extension the total error rate). Both prior work~\citep{raghuram2020detecting} and our empirical study have verified this observation: if we applied
all the said techniques, then many of our in-distribution samples are rejected.\footnote{We like to re-iterate the point we made in Section~\ref{sec:theory} that the definition of OOD (and thus ID) is an elusive concept. Indeed, almost all practical datasets contain so-called outliers for the same reason.} For instance, on MNIST where the classification accuracy on the clean inputs is around $99\%$, a stringent variant of our method can reject $10\%$ of in-distribution samples. This might be prohibitive in some applications. This motivated us to systematically loosen the aforementioned constraints. We based this study off of the established framework of NP-classification and our formulation of the problem in Section~\ref{sec:theory} contains all such considerations. In our empirical study, we found that weighting the \pvalues and effect sizes such that deeper layers are weighted more to balance an effective trade-off. 

The requirement to employ a separate moderately-sized validation set at inference may also not be practical for certain applications and datasets of limited size. In the machine learning literature, usually a smaller fraction of the training set is used for validation and hyper-parameter tuning. However in \pdknn, this assumption ensures that test statistics are independent across the layers. Moreover, it allow us to properly model the decision boundaries found on the training set. In practice, this assumption can also be relaxed (as have been in prior work~\citep{raghuram2020detecting}) which is what we also did in parts of our empirical study.
\section{Conclusions}

We presented \pdknn, a method for detecting out-of-distribution inputs at test time given an already trained neural network classifier. To provide grounded estimates of uncertainty, our approach analyzes internal layer representations with statistical hypothesis testing. Our approach is best studied in a Neyman-Pearson classification setup. We also positioned it with respect to recent work in selective classification with arbitrarily distributed inputs at inference.
Our evaluation on both synthetic and standard datasets shows superior performance to recent proposals in the literature. Among the many directions for future work, we are particularly interested in the benefits of composing the \pvalues associated with the predictions of different neural networks (e.g., when these networks operate on different modalities of the same input).

\section*{Acknowledgements}

We would like to acknowledge our sponsors, who support our research with financial and in-kind contributions: CIFAR through the Canada CIFAR AI Chair program, DARPA through the GARD program, Intel, Meta, NFRF through an Exploration grant, and NSERC through the Discovery Grant and COHESA Strategic Alliance. Resources used in preparing this research were provided, in part, by the Province of Ontario, the Government of Canada through CIFAR, and companies sponsoring the Vector Institute. We would like to thank members of the CleverHans Lab for their feedback.

\appendix 
\newpage

\onecolumn

\section{Additional Information on the Method}

\subsection{Pair-wise vs. One-vs-All Distance Testing}
For a given layer $\lambda \in [L]$, we have the activations per layer for the new test input $\Test_\layer$ and distances of the nearest neighbors $\bdist_{\layer, \class}$ from each class $y$. In the statistical tests, we do \textit{pair-wise} comparison between two classes using the distances $\bdist$ to determine to which class the new input $\test$ belongs to. The alternative approach is \textit{One-vs-All}, where we test distances from a selected class vs distances from all other classes. We observe that the latter method performs worse. As an illustrative example, if $\test$ is close to two classes $A,B$ and very far away from other classes, then these two close classes $A,B$ are accepted as potential classes for $\test$. In such a case, the \textit{pair-wise} comparison between $A,B$ is required to determine the final class.

We can also construct a case, where the \text{pair-wise} comparison is not sufficient. If we consider all the reference (training or calibration) points, then $\test$ can lie on a a very wide manifold from class $A$ but can also be surrounded by a circle-like manifold from class $B$. If we consider distances of all points from class $A$ to $\test$, then they might be larger than the distances from class $B$, even though $\test$ belongs to class $A$. In the multi-dimensional spaces, to alleviate this problem and be able to obtain useful approximation from our distance metric, we have to narrow down the considered space to the vicinity of $\test$ by finding the $k$ nearest neighbors. 

\subsection{Other Possible Inference Methods Based on $p$-values}
The extreme is to remove a class from consideration if it does not appear in each layer. In other words, if a class does not appear in any neighbor across all layers, we discard the class (as a potential correct prediction). We also have a threshold that there have to be at least 10\% neighbors of a given class to consider the class within a layer. If we do not set the significance level, then only 0.4\% examples out of 10000 clean \mnist test samples are rejected. Only these samples have confidence lower than 1. All other samples have the confidence of 1 (and their \pvalues are equal to 0). On the other hand, only 44\% of notmnist samples are rejected and because their confidence is lower than 1 (56\% \pvalues are equal to 0). Thus, this is a too strict/conservative test. We have to consider more \pvalues to obtain higher than 0 \pvalues and these \pvalues should be discriminate enough to distinguish between in distribution and out-of-distribution samples. In the above approach, only 1 out of 1000 samples requires a comparison between 3 classes.

One way to tackle the problem of the too confident predictor is to fill in the missing \pvalues. For example, if class $c1$ is present in all layers while class $c2$ is only in layers 2,3,4 but absent in layer 1, then we set the \pvalue that intuitively denotes the \textit{probability} that class $c2$ is a better prediction than class $c1$ in layer 1 to 0. If classes $c1$ and $c2$ are considered in some layers but both are absent from, e.g., the first layer, then we set the corresponding \pvalues to 1 for both classes in this first layer. This allows us to set the threshold of 5\% acceptance rate for in distribution samples using a significance level of 0.05, however, the acceptance rate of the out-of-distribution samples is about 33\% (the previous method gave only about 19\%).

The \textit{occam} method (with dictionary of dictionaries) can converge to the the \textit{combine} method (a combination of \pvalues in tensors). We have to fill in the missing values using \pvalue 0 or 1. This has to be done, for instance, on the level of layers, where some classes are missing in some layers, and on the level of the class aggregation when there are missing classes from consideration (they do not appear in the nearest neighbors found). The \textit{occam} method has a more complex implementation so we retain the \textit{combine} method as our primary one.

\section{Details for Experiments from Table~\ref{tab:compare}}
\label{sec:details_hyper}

Note, we do not strictly follow the weighted average from~\cite{vovk2019combining} but weight the \pvalues before the (standard) averaging and then multiply by the factor $\min(2, \frac{1}{\max(w)})$, where $w$ is the vector of weights per layer. We do a few simplifications compared to Algorithm~\ref{alg:pdknn} and present the steps in Algorithm~\ref{alg:pdknn-table}.

\begin{algorithm}
    \SetAlgoLined

    \KwInput{Test point: $\bm{\tilde{x}}$, test point activations: $\{\bm{\tilde{x}}_{\lambda}\}_{\lambda=1}^{L}$, activations from the training set: $\{\bm{X}_{\lambda}\}_{\lambda=1}^{L}$, labels from the training set: $\bm{y}$, number of neighbors: $k$, significance level: $\alpha$.}

    \KwOutput{Classification $\tilde{y} \in [C]$ or abstain $\bot$}
        
     \For{$\lambda \in [L]$}{
     
        $\bm{\mu} \longleftarrow $ \texttt{get\_distances($\bm{X}_{\lambda}, \bm{\tilde{x}}_{\lambda}$)}\;
      
        $\bm{\hat{\mu}}, \bm{indices} \longleftarrow $ \texttt{get\_smallest($\bm{\mu}, k$)}\;
      
        $\bm{\hat{y}} \longleftarrow $ \texttt{get\_classes($\bm{y}, \bm{indices}$)}\;
        
        $ \mathcal{P}_\lambda \longleftarrow \texttt{pairwise\_t($\bm{\mu}, \bm{\hat{y}}$)} $\;
     }
     
     $\bm{p} \longleftarrow \texttt{aggregate\_p($\mathcal{P}$)} $\;
     
     $\tilde{p} \longleftarrow \texttt{min($\bm{p}$)}$\;
     
     $\tilde{y} \longleftarrow \texttt{argmin($\bm{p}$)}$\;
     
     \If{$\tilde{p} < \alpha$}{
        \Return{$\tilde{y}$}
     }
     
     \Return{$\bot$}
     
     \caption{\pdknn prediction algorithm for results from~Table~\ref{tab:compare}.}
     \label{alg:pdknn-table}
    \end{algorithm}

For results in Table~\ref{tab:compare}, we present the classification accuracy for the datasets and methods in Table~\ref{tab:compare-acc} and the parameters used in Tables~\ref{tab:compare-params1} and~\ref{tab:compare-params2}.

We follow the setup from~\citet{dknn} and use the same deep network architecture (as described there in Table III) as well as the parameters for FGSM and PGD (BIM) attacks (Table II). 

\begin{table}
\caption{\textbf{Classification accuracy} of the models from Table~\ref{tab:compare} on the entire test set.
}
\label{tab:compare-acc}
\vskip -0.2in
\begin{center}
\begin{sc}
 \begin{tabular}{ l c c c c } 
 \toprule
  Model & \mnist & \fashion & SVHN & GTSRB \\
 \midrule
 NN & .988 & .905 & .902 & .928 \\
 \dknn & .988 & .897 & .904 & .930 \\
\pdknn & .991 & .907 & .901 & .914 \\
\bottomrule
\end{tabular}
\end{sc}
\end{center}
\vskip -0.1in
\end{table}

\begin{table}
\caption{\textbf{Parameters} for the classifiers from Table~\ref{tab:compare} for MNIST and FashionMNIST. $\tau$ is the confidence threshold, $k$ is number of nearest neighbors, $w$ are weights per layer.
}
\label{tab:compare-params1}
\vskip -0.2in
\begin{center}
\begin{sc}
 \begin{tabular}{ l c c c c c c } 
 \toprule
  \multicolumn{1}{c}{} & \multicolumn{3}{c}{\mnist} & \multicolumn{3}{c}{\fashion} \\
   \cmidrule(r){2-4}
   \cmidrule(r){5-7}
 \textbf{Model} & \textbf{$\tau$} & \textbf{$k$} & \textbf{$w$} & \textbf{$\tau$} & \textbf{$k$} & \textbf{$w$} \\ [0.5ex] 
 \midrule
 NN & .974 & N/A & N/A & .952 & N/A & N/A \\
 \dknn & .05 & 75 & N/A & .11 & 75 & N/A\\
\pdknn & .0019 & 100 & .02 .02 .02 .94 & .025 & 100 & .1 .1 .1 .7\\
\bottomrule
\end{tabular}
\end{sc}
\end{center}
\vskip -0.1in
\end{table}

\begin{table}
\caption{\textbf{Parameters} for the classifiers from Table~\ref{tab:compare} for SVHN and GTSRB. $\tau$ is the confidence threshold, $k$ is number of nearest neighbors, $w$ are weights per layer.
}
\label{tab:compare-params2}
\vskip -0.2in
\begin{center}
\begin{sc}
 \begin{tabular}{ l c c c c c c } 
 \toprule
  \multicolumn{1}{c}{} & \multicolumn{3}{c}{SVHN} & \multicolumn{3}{c}{GTSRB} \\
     \cmidrule(r){2-4}
   \cmidrule(r){5-7}
 \textbf{Model} & 
 \textbf{$\tau$} & 
 \textbf{$k$} & 
 \textbf{$w$} & 
 \textbf{$\tau$} & 
 \textbf{$k$} & 
 \textbf{$w$} \\ [0.5ex] 
\midrule
 NN & .672 & N/A & N/A & .870 & N/A & N/A\\
 \dknn & .08 & 75 & N/A & .04 & 75 & N/A \\
\pdknn & .15 & 100& .05 .07 .07 .81 & .058 & 100 & .05 .05 .1 .1 .7 \\
\bottomrule
\end{tabular}
\end{sc}
\end{center}
\vskip -0.1in
\end{table}

\begin{table}
\caption{\textbf{Average confidence $\mu$} for the classifiers from Table~\ref{tab:compare}.
}
\label{tab:compare-params}
\vskip -0.2in
\begin{center}
\begin{sc}
 \begin{tabular}{l l c c c c } 
 \toprule
  \multicolumn{1}{c}{Attack} & \multicolumn{1}{c}{Model} & \multicolumn{1}{c}{\mnist} & \multicolumn{1}{c}{\fashion} & \multicolumn{1}{c}{SVHN} & \multicolumn{1}{c}{GTSRB} \\
 \midrule
 \multirow{3}{*}{None} & NN & .99 & .977 & .921 & .958 \\
 & \dknn & .663 & .552 & .451 & .460 \\
  & \pdknn & .991 & .969 & .953 & .955 \\
  \midrule
 \multirow{3}{*}{OOD1} & NN & .942 & .952 & .688 & .549  \\
 & \dknn & .063 & .142 & .088 & .066  \\
  & \pdknn & .937 & .941 & .783 & .821 \\
   \midrule
 \multirow{3}{*}{OOD2} & NN & .607 & .977 & .700 & .597  \\
 & \dknn & .046 & .552 & .017 & .073  \\
  & \pdknn & .943 & .969 & .787 & .820 \\
\midrule
\multirow{3}{*}{Feat} & NN & .0 & .850 & .788 & .677 \\
 & \dknn & .071 & .191 & .170 & .117 \\
 & \pdknn & .754 & .944 & .849 & .865 \\
\midrule
\multirow{3}{*}{FGSM} & NN & .929 & .984 & .867 & .765 \\
 & \dknn & .073 & .205 & .125 & .131 \\
     & \pdknn & .956 & .942 & .826 & .861 \\
 \midrule
 \multirow{3}{*}{PGD} & NN & .999 & 1.0 & 1.0 & 1.0 \\
 & \dknn & .141 & .219 & .378 & .352 \\
& \pdknn & .975 & .948 & .957 & .943 \\
 \midrule
  \multirow{3}{*}{$\measuredangle$45} & NN & .837 & .914 & .693 & .813 \\
 & \dknn & .166 & .093 & .061 & .156 \\
 & \pdknn & .972 & .933 & .764 & .877 \\
\bottomrule
\end{tabular}
\end{sc}
\end{center}
\vskip -0.1in
\end{table}

\section{Potential Negative Societal Impacts of Our Work}

Our work aims at improving the inference of the already trained neural network classifiers. Thus, we focus on a general-purpose methodology and not specific applications that could have more direct potential impact on society. On the other hand, we can imagine that our method could be used to train a classifier that refrains from providing answers for specific ethnic groups and classifying them as OOD. To mitigate such a harm, the train and validation sets should include samples for all possible ethnic groups. 

\end{document}